\documentclass[preprint,1p,times]{elsarticle}
\usepackage{amssymb}
\usepackage{subfigure}
\usepackage{graphicx}
\usepackage{amssymb,booktabs}
\usepackage{multirow}
 \usepackage[table,xcdraw]{xcolor}
\usepackage{amsmath}
\usepackage{wrapfig}
\usepackage{mathtools}
\usepackage{algorithm}
\usepackage{algpseudocode}
\usepackage{lipsum}
\makeatletter
\def\BState{\State\hskip-\ALG@thistlm}
\makeatother
\algnewcommand\algorithmicinput{\textbf{Input:}}
\algnewcommand\INPUT{\item[\algorithmicinput]}
\algnewcommand\algorithmicoutput{\textbf{Output:}}
\algnewcommand\OUTPUT{\item[\algorithmicoutput]}
\algnewcommand\algorithmicprocedurea{\textbf{Procedure:}}
\algnewcommand\PROCEDUREA{\item[\algorithmicprocedurea]} 
\algnewcommand\algendproc{\textbf{end~procedure}}
\algnewcommand\ENDPROCEDURE{\item[\algendproc]}
\algrenewcommand{\algorithmicforall}{\textbf{for each}}

\usepackage{eqparbox}
\journal{Signal Processing: Image Communication}

\begin{document}
\begin{frontmatter}
\title{A Classifier-guided Approach for Top-down Salient Object Detection }
\author{Hisham Cholakkal,  
         Jubin Johnson,
          Deepu Rajan
}
%
%
\address{ 
              School of Computer Engineering, \\    
              Nanyang Technological University, Singapore\\      
              } 

\begin{abstract}
 We propose a framework for top-down salient object  detection that incorporates a tightly coupled image classification module.  
  The classifier is trained on novel category-aware sparse codes computed on  object dictionaries  used for saliency modeling. 
    A misclassification indicates that the corresponding saliency model is inaccurate. Hence, the classifier selects images for which the saliency models need to be updated. The category-aware sparse coding  produces better  image classification accuracy as compared to conventional sparse coding with a reduced computational complexity. 
  A saliency-weighted max-pooling is proposed to improve image classification, which is  further used to refine the saliency maps.
 Experimental results on Graz-02 and PASCAL VOC-07 datasets demonstrate the effectiveness of salient object detection. Although the role of the classifier is to support salient object detection, we evaluate its performance in image classification and also illustrate the utility of thresholded saliency maps for image segmentation.
\end{abstract}
\begin{keyword}
Saliency \sep Top-down saliency \sep Image classification\sep Semantic segmentation
\end{keyword}
\end{frontmatter}
%
%

\section{Introduction}

Visual attention in humans is largely affected by past experiences of the visual world \cite{BookSaliency}. This phenomenon, includes \textit{contextual cuing} which helps human brain to respond faster  by spending less neural resources at spatial locations where the probability of a target is low.   Mimicking contextual cuing, top-down saliency approaches learn the relevance of image features and spatial locations from a set of training images 
which helps to reduce the search space for computationally expensive tasks such as object segmentation and detection. On the other hand, in bottom-up saliency detection, visual rarity is used to identify salient regions and hence, it often fails in the presence of background clutter due to which the salient object does not 'pop-out'.

%

 In general, there are two types of top-down saliency detection approaches-- knowledge-based and task-based~\cite{BookSaliency}.  Knowledge-based approaches aim to learn the prior knowledge about  relevant objects under free-viewing conditions \cite{li2014visual, Objt_sal_category_indep,itti2001feature,ZhaoKoch2011,Navalpakkam2007}. 
  e.g., attention to faces. 
 Task-based top-down saliency models use prior knowledge to infer a probability map that suppresses image regions not relevant to the task \cite{topdownCVPR2012,topdownDSD,Torralba_topdown,GauravCVPR2012}. 
 Our approach falls under the second category, where the task is estimating the presence of objects from a particular category in an image (image classification) as well as identifying the image locations of those objects (object localization). 
The regions that are closer to the pre-learnt priors are assigned higher saliency values, even if they are not salient in the 'pop-out' sense.

 Salient object detection is different from object detection in that the latter  aims to produce  a tight rectangular bounding box around the object of interest. 
        Even for a moderately sized image, the algorithm needs to  sample and classify millions of rectangular boxes with arbitrary shapes, sizes or locations \cite{CVPR11_0254}, which is computation intensive.
       In order to reduce  computations, recent object detection approaches\cite{R-CNN} use bottom-up saliency \cite{objectness} to extract regions of probable objects in an image. Since these region proposals are category-independent, they still result in a large number of  rectangular boxes that are irrelevant to the task. Task-specific top-down saliency approaches, such as the one proposed, can reduce computations significantly by extracting few task-relevant image regions.        
        Being a pre-processor, their training and inference is faster compared to dedicated object detectors, and hence often computed with coarse spatial granularity.

  Object localization is often used as a synonym for object detection\cite{cinbis2014multi}.  However  other approaches \cite{shapeMaskIJCV2012, fulkerson2008localizing} use the term  to represent the process of identifying probable object locations, that need not be a pixel-accurate estimation as in object segmentation, but more accurate than a rectangular box estimated by object detection.  In this paper we use the term 'object localization' to represent probable object locations. Our saliency map, which is a probability map that peaks at locations of a task defined object, can be used for object detection and segmentation applications. 

\begin{figure}[t!]
 \centering
  \includegraphics[scale=0.6, clip=true, trim=2.2cm 13.3cm 12.2cm 2.5cm]{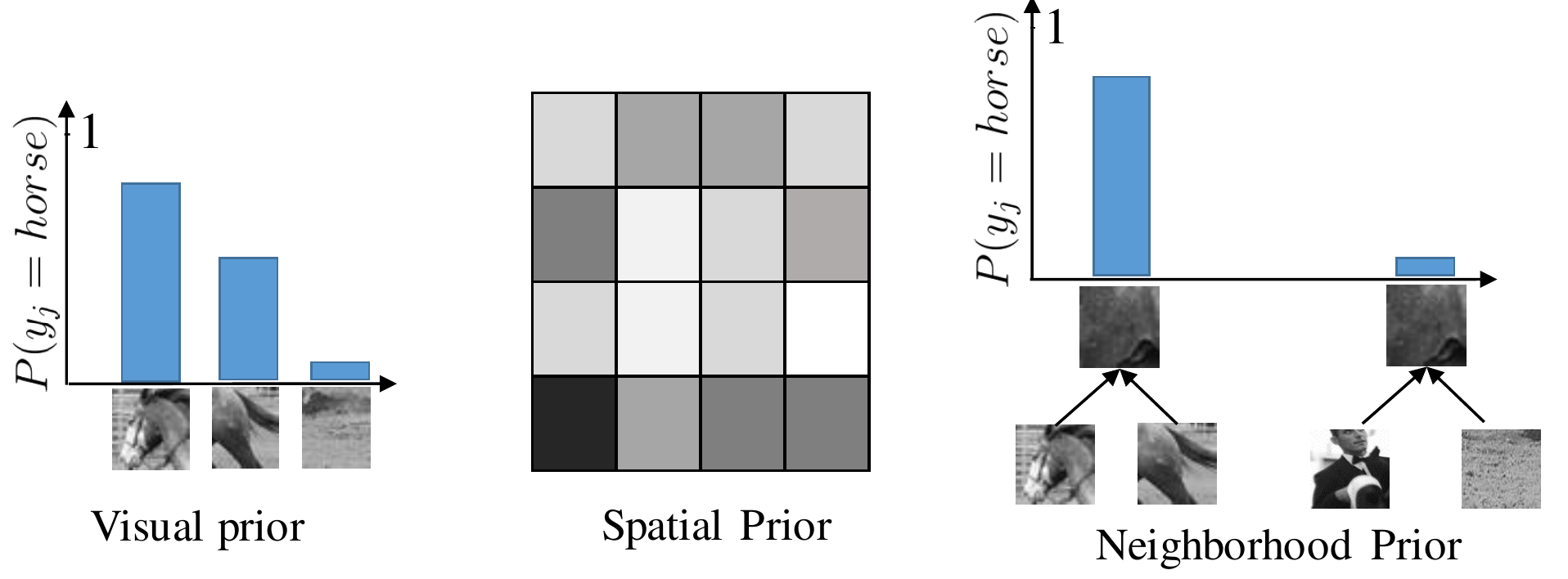}\\
  \hspace*{0.1cm}(a)\\
  
    \includegraphics[scale=0.4, clip=true, trim=0.0cm 16.7cm 0cm 0.5cm]{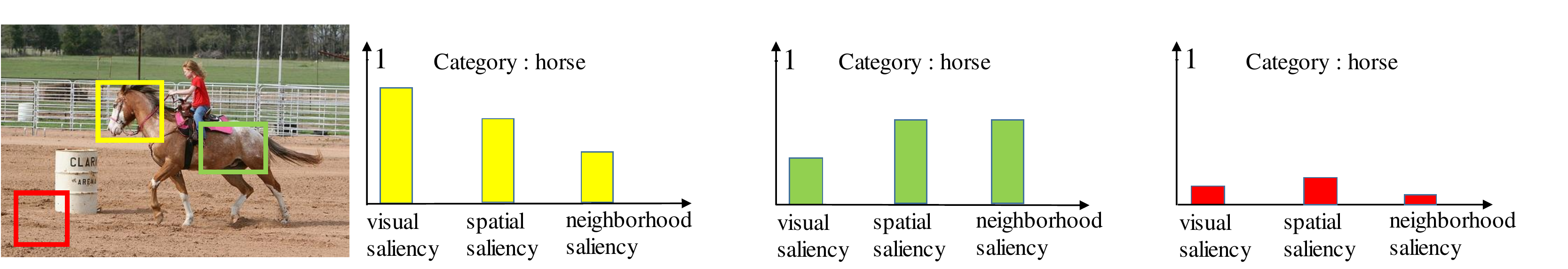}\\
\hspace*{0.1cm}(b)\\
 \caption{Priors for task-specific top-down saliency computation. (a) task-specific priors for horse category learnt from the training images.  The visual, spatial and neighborhood saliency values  for yellow, green and red colored image boxes in (b) are shown in their respective colors.  The spatial prior is a 2-D distribution of horse patches in a $4\times 4$ spatial grid with white indicating high probability and black indicating low probability. The  horse's head (yellow  box)  has high correlation to the horse visual prior, resulting in large visual saliency.  Similarly, it is less likely to find a horse patch at the position of the red box, resulting in lower spatial saliency. }
 \label{fig:FigTopDownPriorL}
 \end{figure}

In the proposed method, feature priors suited for the task of joint object localization and image classification, namely visual, spatial and neighborhood priors are used to compute saliency.  We learn the visual and neighborhood priors  based on SIFT~\cite{SIFT_Lowe} features through a  conditional random field (CRF)~\cite{topdownCVPR2012}, and the spatial prior through a spatial pyramid \cite{ScSPM_CVPR09}. 
%
Fig.\ref{fig:FigTopDownPriorL} illustrates the idea of task-specific top-down salient object detection.  
Suppose the visual, spatial and neighborhood prior probabilities of features learnt for horse category are as shown in Fig.\ref{fig:FigTopDownPriorL} (a). This prior knowledge is used to estimate  visual, spatial and neighborhood saliency separately  for an image as shown in  Fig.\ref{fig:FigTopDownPriorL} (b). The saliency values for yellow, green and red colored boxes are shown as bars with their corresponding colors. If  a particular feature in a box has  high prior probability of belonging to the horse category, then the box is assigned a high saliency.
 The  yellow colored box  is  assigned with high visual saliency due to the high visual prior of the horse's head. 
  
\subsection{Combining top-down saliency and image classification}
               	Based on the task at hand, task-specific top-down saliency approaches can be  grouped into  approaches for (i) image classification \cite{GauravCVPR2012,topdownDSD,colorICCV2009} and  those for (ii)  object localization \cite{Torralba_topdown,topdownCVPR2012,topdownBMVC2014}. In this paper, we demonstrate that the proposed saliency model  is useful for both image classification and object localization tasks. 
                               The saliency model of \cite{topdownCVPR2012} uses   neighborhood and visual priors but lacks a spatial prior. On the other hand, the image classification approach of \cite{ScSPM_CVPR09} has multi-scale spatial information, but lacks neighborhood prior.        
                                   This paper is motivated by a need for  top-down salient object detection and image classification joint framework that uses spatial, neighborhood and visual priors. 
               A sparse coded spatial pyramid max-pooling (ScSPM) image classification can be improved by improving the discriminative quality of the dictionary, which can be obtained from the dictionary update of \cite{topdownCVPR2012}. Similarly, saliency maps can be improved using the image classifier by leveraging  information about presence of the object  in an image. 
                        Thus, we propose an  interconnected and mutually benefiting saliency-classification framework  that helps reduce  the computational cost.\\ 
                                                  Next, we briefly describe the main processes in our joint framework.
               

                 

       \textit{ Category-aware sparse coding.}
                              ScSPM-based image classifiers are known to perform better~\cite{DevilInDetail} if the sparse codes are computed on a global dictionary whose atoms are representative features from all categories. Since the saliency models are developed based on  sparse codes  using individual object dictionaries, it would be computationally expensive to form a global dictionary using k-means clustering of thousands of patches from all categories and recompute sparse codes with respect to that dictionary. The proposed category-aware sparse coding  utilizes the discriminative dictionaries  already learned during saliency modeling, thereby enabling tight coupling between saliency models and the image classifier. This strategy helps in reducing the computational cost of feature coding for the classifier, while improving its classification accuracy. 

          \textit{Classifier-guided saliency model training.} The image classifier used to update the saliency model is trained using a training set  and validated using a validation set. The misclassified images during validation indicate that the features of those images are not represented adequately in the discriminative dictionary. The top-down saliency module is updated using such images. This classifier-guided saliency model training helps to improve not only the top-down saliency component, but also the image classification accuracy.                               

         \textit{ Saliency-weighted max-pooling.} Conventional ScSPM image classifier~\cite{ScSPM_CVPR09} is blind to  max-pooled vectors from  the spatial pyramid blocks that contain an object and from those that do not contain  any object. Hence, ScSPM often performs poorly in the presence of high background clutter. A novel  saliency-weighted max-pooling proposed in this paper helps improve the performance.        
                   When a saliency model of a particular class is applied to an image, it highlights those patches that are likely to contain object parts belonging to that class. 
          High saliency values appearing in a negative training image indicate false positive patches. Weighting corresponding max-pooled sparse codes with this high value and training the SVM~\cite{liblinear}  with a negative label will help the image  classifier to label similar test images as negative. 
                                   
     \textit{Saliency refinement.}     The  top-down saliency approaches of \cite{topdownCVPR2012,topdownBMVC2014}  compute saliency of an image region
                without ascertaining whether the target object is present or not.
                 Due to this, they  often end up in producing false detection on negative images, which reduces the accuracy.          
                                In our framework, the image classifier is used to quantify the likelihood of the presence of an object and to refine the saliency map using a novel saliency refinement.

%

                  \begin{figure}[t!]
          \centering
           \includegraphics[scale=0.47, clip=true, trim=0.23cm 4.0cm 5.2cm 0.5cm]{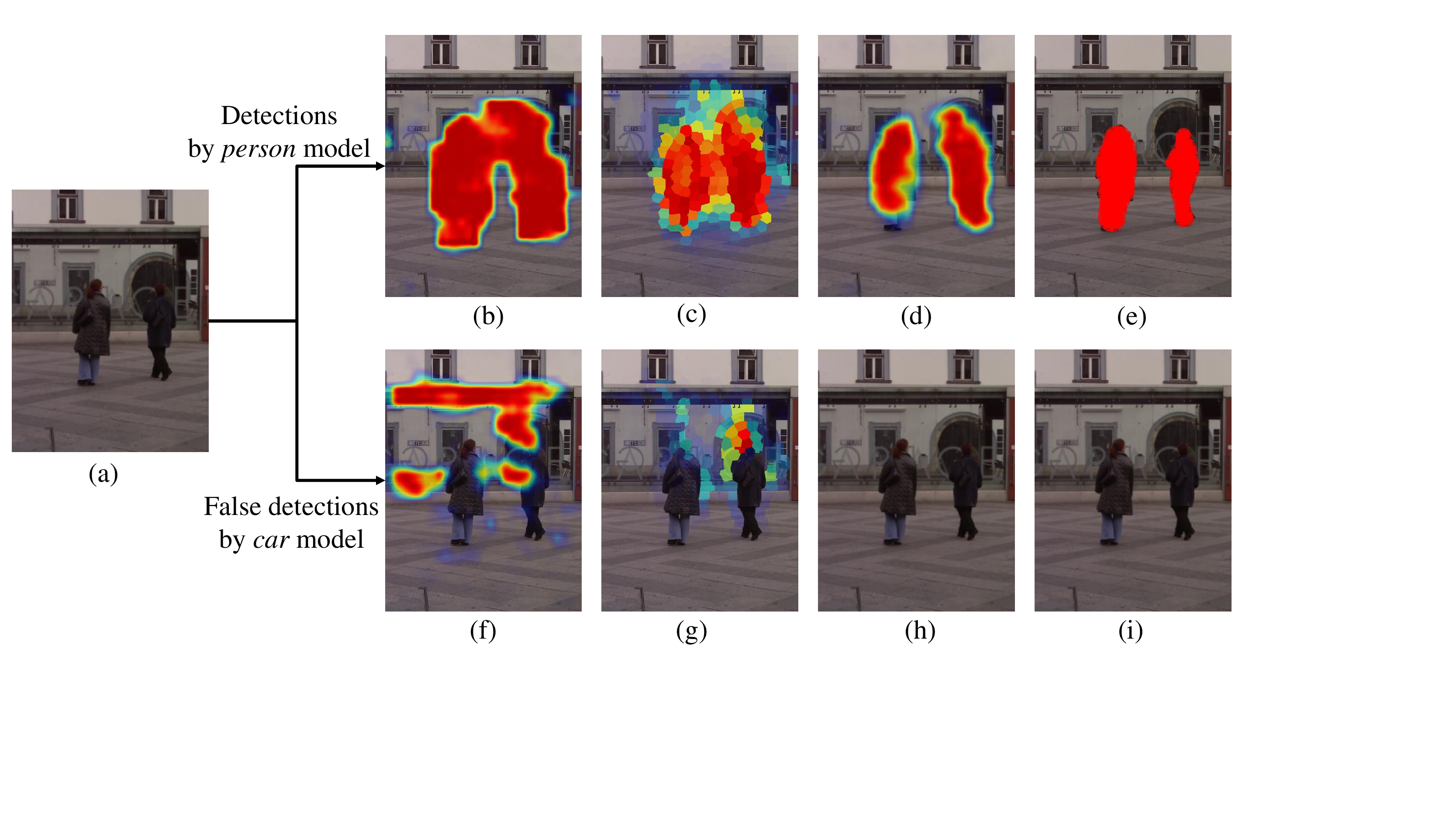}\\
          \caption{Comparison of saliency maps for an (a)~input image by (b)~Yang and Yang~\cite{topdownCVPR2012}, (c)~Kocak~\textit{et al.}~\cite{topdownBMVC2014},  (d) proposed method and (e)~ground truth on \emph{person} model. False detections on applying \emph{car} model of (f)~Yang and Yang~\cite{topdownCVPR2012} and (g)~Kocak~\textit{et al.}~\cite{topdownBMVC2014} are eliminated by (h) the proposed method resulting in a saliency map that is visually similar to (i) the ground truth.}
          \label{fig:saliencyComparison}
          \vspace*{-0.45cm}
          \end{figure}
        A preview of the effectiveness of our joint framework for top down saliency and image classification for the task of salient object detection is shown in Fig.~\ref{fig:saliencyComparison}.
         Fig.~\ref{fig:saliencyComparison}(a) shows an image containing two persons in a cluttered background.
         The person model of~\cite{topdownCVPR2012} and \cite{topdownBMVC2014} are unable to distinguish between the  persons due to their proximity in the image as seen in Fig.~\ref{fig:saliencyComparison}(b) and (c) respectively. Our framework produces a saliency map (Fig.~\ref{fig:saliencyComparison}(d)) closer to that of ground truth (Fig.~\ref{fig:saliencyComparison}(e)). Fig.~\ref{fig:saliencyComparison}(f) and (g) show the false detections by  \cite{topdownCVPR2012} and \cite{topdownBMVC2014} respectively when their car saliency models are applied on the input image. As seen in Fig.~\ref{fig:saliencyComparison}(h), with the help of the classifier in our framework, we avoid these false detections.
         
     In summary, the key idea of the proposed method is to add the image classification module both to update the saliency models, and to refine the saliency map for a given test image. On the other hand, the saliency inferred from the saliency models can also be used to improve the accuracy of the image classifier.  The major contributions of this paper are the following:
         \begin{enumerate}
         \item A novel category-aware sparse coding strategy, which is computationally more efficient as compared to conventional sparse coding. 
         \item A novel framework to train saliency models with the help of a classifier.
         \item A novel saliency-weighted image classification framework.
         \item Refinement of saliency maps using image classifier confidence.
         \end{enumerate}

\section{Related Work}
First, we review the related works in task-specific top-down saliency estimation and image classification, which are the two major components of the proposed joint framework. 
Since the usefulness of the saliency map obtained by the proposed approach is demonstrated
for image segmentation and localization applications we briefly review algorithms for those applications also.
\subsection{Task-specific top-down saliency }

 As mentioned in the previous section, task-specific top-down saliency approaches  can be broadly categorized  into approaches for (i) object localization  and those for (ii) image classification.
\subsubsection{Task-specific top-down saliency approaches for object localization}
                     
                       
                     
                    
                       
                     The locality-constrained linear coding (LLC)  of  SIFT features followed by max-pooling in a spatial context is used for saliency estimation in \cite{topdown_Contextual}. 
                        Even though  they use the neighborhood prior through contextual max-pooling, it lacks a spatial prior.
                         Khan and Tappan~\cite{icip2013discriminative} use label and location-dependent smoothness constraint in a sparse code formulation to improve the pixel-level accuracy compared to the conventional sparse coding, but with additional computation cost. 
                     %

                        Neighborhood prior of image patches on CRF  is used for saliency estimation in \cite{topdownCVPR2012}. The joint learning of CRF weights and dictionary was observed to improve the accuracy of the saliency map only marginally after 10 cycles, which makes the further  learning of dictionary computationally less efficient. Moreover, this approach has limited ability to discriminate among the object categories. 
                     \cite{topdownBMVC2014} improves on \cite{topdownCVPR2012} by replacing patch-level SIFT features with RGB-based features from superpixels and by incorporating objectness~\cite{objectness}. Although this improved the accuracy of distinguishing objects from background, its ability to discriminate between object categories did not improve, causing large number of false detections if the test image contained irrelevant objects. By coupling saliency prediction with classification, such false detections are avoided in our framework. Also, using an image classifier to select the training images for  \cite{topdownCVPR2012}, we could improve the saliency models, even with reduced training time. 
                       \subsubsection{Task-specific top-down saliency approaches for image classification}
                       
                     The appearance statistics of discriminative features from a pre-defined filter bank are used in \cite{topdownDSD}  to distinguish different object classes. These salient features are learned using a weakly supervised approach, which are used for image classification. However, by only considering the image-level statistics of a feature and not its neighborhood information, they are unable to remove background patches leading to poor performance on datasets with heavy background clutter and viewpoint variations.
                       
                              In \cite{GauravCVPR2012}, saliency is used to determine  the discriminative patches  for image classification.
                               Saliency of a local region is calculated by considering its appearance along with its spatial location. Integrated max-margin learning is used to learn saliency and classifier.
                              In \cite{CVPR11_0254}, a random forest with discriminative decision tree is used to mine out the discriminative patches for fine-grained image classification.
                        Category-specific color features  are used in   \cite{colorICCV2009, colorAttentionForObjectRecognition} to modulate SIFT features. Shape features from the regions with higher color-attention are given more weight than those from lower attention.
												In \cite{topdownCategINRIA}, which is closest to our approach, a classifier is learned using randomly selected, random sized sub-windows from an image, which is then used to build and update a saliency map. Using this saliency map, the classifier samples  more sub-windows in the salient region. The drawback here is that an error in the initial estimate of the saliency map gets propagated to consecutive iterations resulting in the failure of both  classifier and saliency estimation. To avoid this, in our framework, training of top-down saliency model is separated from the image classifier and they are integrated only after saliency reaches a certain level of accuracy.
                       
                       \subsection{Classification}
                      Spatial Pyramid Matching (SPM)~\cite{SPM_CVPR_06} has been widely used in image classification. In sparse coded SPM (ScSPM)~\cite{ScSPM_CVPR09}, features were sparse coded and then max-pooled. This reduced the classifier complexity to $O(N)$ compared to $O(N^{2}\  or \  N^{3})$ in SPM for $N$ training images. Since max-pooled vectors from  the spatial pyramid blocks that contain an object and those that do not contain  any object are considered equally, ScSPM often performs poorly in the presence of high background clutter.  We propose saliency-weighted max-pooling to improve this performance. 
                       The computational complexity of ScSPM is further reduced in LLC~\cite{LLC_CVPR10} by imposing  feature locality constraint, and in~\cite{ProductSparseCoding_CVPR14} by compromising on the accuracy through division of the dictionary into cartesian product of sub-dictionaries. 
                        In \cite{Vinay_IJCV}, feature extraction from an automatically selected  bounding box  around  objects is shown to improve image classification accuracy. 
                        However, this method needs iterative expansion of latent parameter space for effective localization of the object, which is computationally expensive.

                      \subsection{Related Applications : Object localization and segmentation}
                     
                      In ~\cite{shapeMaskIJCV2012}, bag-of-words representation is improved by spatial weighting of features using shape masks, causing  foreground features to be boosted, thereby decreasing the influence of background clutter. High dimensional hypothesis clustering of shape mask is used for localization which requires separate annotation for each object within an image. i.e, if multiple objects from same category are present in an image, each of them needs to be labeled separately. Additionally, training the model requires images to be marked as \emph{difficult} or \emph{truncated}. The proposed method produces better localization result compared to~\cite{shapeMaskIJCV2012} without the added requirement of such annotations. 
                      
                      Bag-of-feature based classification  of  local image regions having a fixed size is used in~\cite{fulkerson2008localizing}. Absence of spatial consistency and context-type constraints causes many false detections leading to less accurate  localization of objects.
                     The class segmentation approach  of~\cite{fulkerson2009class} uses superpixels as the basic unit to build a classifier using histogram of local features within each superpixel. Histograms in a neighborhood are aggregated and used to regularize the classifier. A CRF  built on superpixel graph further improved the segmentation accuracy. The performance of this model  depends on the neighborhood size whose optimum size varies across object categories. 
                     
                      A fast and robust object segmentation proposed in \cite{aldavert2010fast}  computes multi-class pixel-level object segmentation of an image through an integral linear classifier built on bag-of-words representation of local feature descriptors. Here, a large dictionary of 500,000 words is required and they use a cascaded classifier containing 2 or 3 linear classifiers.
                     \cite{globalBOFinCRFSeg2011} uses bag-of-features for joint categorization and segmentation. The interaction between pixels and superpixels is modeled using random fields and global representation for categorization is achieved through a bag-of-features representation. The number of parameters to be learned for classification increases with increasing number of training images and number of categories. So, it is computationally infeasible to use this approach for large datasets like PASCAL VOC-07~\cite{dataset_pascal-voc-2007}. 
                     The joint categorization and segmentation model proposed in~\cite{JointSegCateg2012jain} simultaneously learns segmentation, categorization, and
                      dictionary learning parameters. Simplicity of bag-of-features representation limits the performance of both.

\section{The proposed joint framework}
\subsection{Brief review on top-down saliency of \cite{topdownCVPR2012}}
 In \cite{topdownCVPR2012}, dense SIFT features are extracted from regular, rectangular gray-scale image patches and their sparse representations are initially computed with a dictionary $D_n$ formed by k-means clustering. The dictionary is formed from cluster centroids and it  represents the most representative patches of an object category. The use of sparse coding of SIFT features results in a more compact and discriminative representation which helps to model feature selectivity for saliency map.
  Using these sparse codes as latent variables, a conditional random field is learned.  
 The CRF node weight $w_n$ is initialized with a binary SVM classifier weight learned on these sparse codes and pairwise energy is set to 1. The dictionary and CRF weights are jointly learned in 20 iterations using max-margin framework. Finally, loopy belief propagation is used to infer saliency values on test images. In the proposed method, the number of iterations is set to 10 to save computations because  the improvement in accuracy  of the model after 10 iterations   is insignificant  compared to its computation cost.
      
     One major drawback of~\cite{topdownCVPR2012} is that its focus  is on distinguishing objects from background and not from other objects, resulting in a large number of false positives as shown in Fig.~\ref{fig:saliencyComparison}(f). Our method overcomes this limitation by integrating  a classifier that is trained on novel category-aware sparse codes, which also results in considerable savings in computation, especially  when dealing with large datasets. %
 \subsection{Brief review on ScSPM image classification \cite{ScSPM_CVPR09} }
  In a ScSPM-based classifier \cite{ScSPM_CVPR09}, dense SIFT features are extracted from gray-scale image patches. SIFT features from training images are used to form a global dictionary. The SIFT features  of an image are sparse coded using this dictionary. The spatial distribution of the features in the image is encoded in the max-pooled image vector  through a multi-scale max-pooling operation of the sparse codes on a 3-level spatial pyramid~\cite{SPM_CVPR_06}. The max pooled feature   is more robust to local transformations than mean statistics in histogram. Biophysical evidence in visual  cortex (V1) \cite{maxPoolBio} also establishes the use of max-pooling.  Image-label pairs of  training images are used to train a linear binary SVM classifier.
  
    
  
  \subsection{System Overview}
 \label{sec:overview}
 \begin{figure*}[t]
      \centering
      \includegraphics[width=1\textwidth, clip=true, trim=2.0cm 3.5cm 6.3cm 0cm]{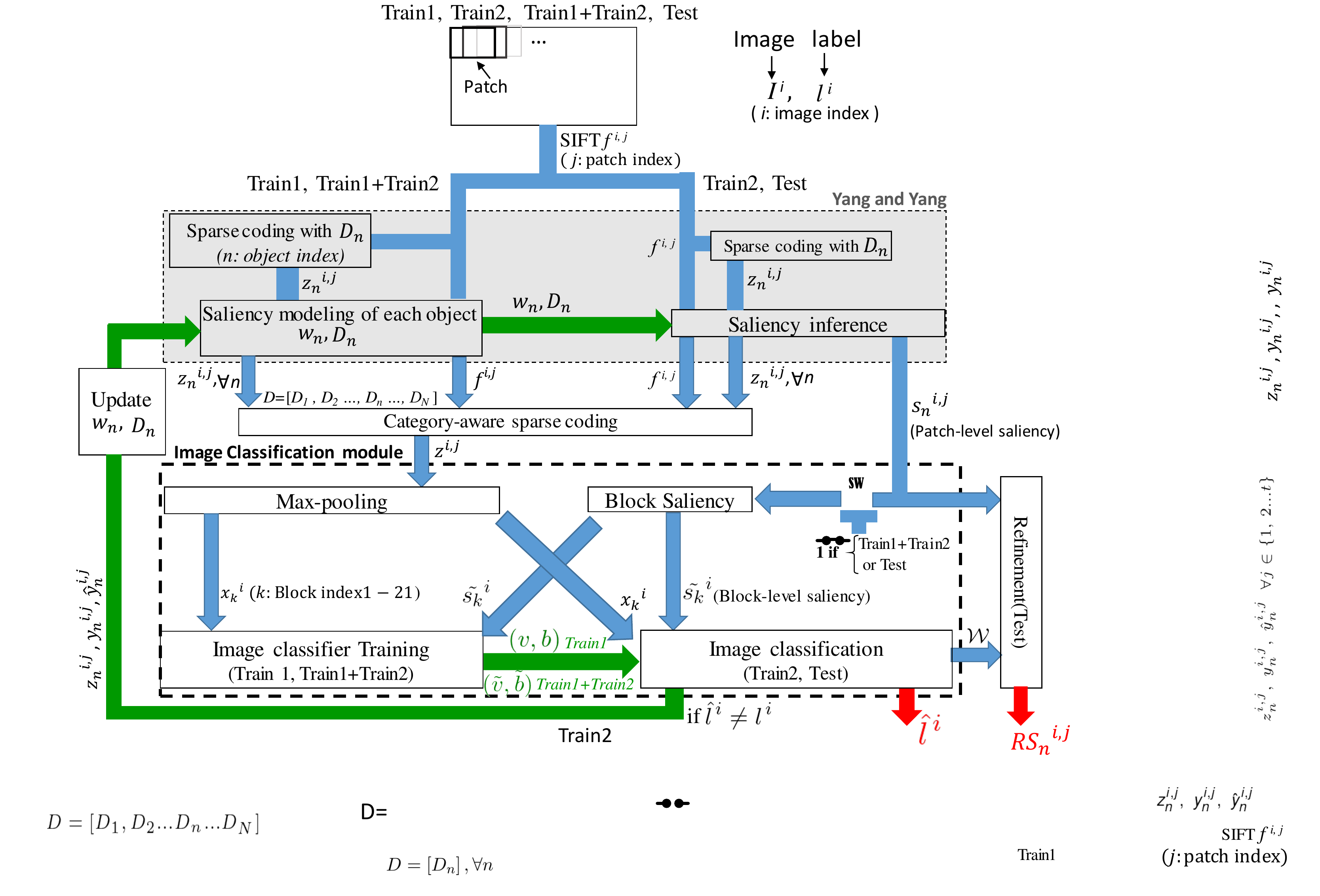}
             \caption{Overview of the proposed framework for classifier-guided salient object detection. The shaded region represents the framework of \cite{topdownCVPR2012}. \textbf{SW} is connected  only during  training and testing of saliency-weighted classifier. \textbf{SW}=1 and \textbf{SW}=0 indicate that the switch is connected and disconnected, respectively. Green arrows indicate output of training stages and red  arrows indicate the final saliency map and classifier output on a test image.}
      \label{JointFramework}
      \end{figure*}

 Fig.~\ref{JointFramework} shows an overview of the proposed joint framework  for salient object detection and image classification. Similar to  the original framework of \cite{topdownCVPR2012} and \cite{ScSPM_CVPR09}, we  only use SIFT features extracted from gray-scale images.
  From every image $I^i$, image patches with a fixed size and grid spacing are extracted. For each patch $j$, its dense SIFT feature ${f}^{i,j}$ and  ground-truth ${y}^{i,j}_n \in \{\text{-1},1\} $ are computed, where -$1$ and $1$ denote  the absence  or presence, respectively, of object $n$ in  patch $j$. 
      We split the training images (Train1+Train2)  into Train1 (training set) and Train2 (validation set), to save computations by avoiding the update of  saliency model with a training image whose features are already well represented in the saliency model.  The discriminative  dictionary $D_n$ for each object category $n$ is initialized with the  centroids of  the clusters formed by k-means clustering  applied on positive SIFT features. 

     Similar to \cite{topdownCVPR2012}, the  sparse codes of SIFT features with  $D_n$ are used as  latent variables in our saliency models. The sparse codes of  positive and negative patches from Train1 images are used to learn a linear SVM and  the SVM weights are used to initialize the the CRF node weight $w_n$. The pair-wise  energy is set to 1 as in \cite{topdownCVPR2012}. Following \cite{topdownCVPR2012}, for each category $n$, the  dictionary $D_n$ and the CRF weight $w_n$ are jointly updated using   Train1 images for 10 iterations.  The updated dictionary $D_n$ represents the most representative patches of the $n^{th}$ object category,  and the CRF weights learnt on sparse codes represents our saliency model.        
Since sparse code is used as a latent variable for saliency modeling, the sparse code ${z}^{i,j}_n$ of ${f}^{i,j}$  with $D_n,~ n\in \{1,\,2,...,\,N\}$, is recomputed in each of these 10 iterations. 

Following  10 iterations, the \textit{ classifier-guided saliency model update} process is iterated 2 times. 
An image classifier that uses $D_n$ and ~${z}^{i,j}_n$ of all $N$ object categories, and trained on Train1 images  is used to  choose  images from Train2  to update saliency models.  
The image classifier performs better if the sparse representations are on a global dictionary $D$ than $N$ separate sparse representations on $N$   discriminative dictionaries $D_n$\cite{DevilInDetail}. 
    So, we propose a \textit{ category-aware sparse coding} strategy that reuses each $D_n$ and ${z}^{i,j}_n$ computed during saliency estimation for $N$ categories to produce a global representation ${z}^{i,j}$ of ${f}^{i,j}$. This  category-aware sparse representation helps to reduce the additional computational requirement for image classification compared to the conventional sparse coding with $D$ \cite{ScSPM_CVPR09}.

 The category-aware representation of all patches in  image $i$ are max-pooled over a multi-scale spatial pyramid and the max-pooled vector in each block $k$ of the spatial pyramid is represented  as ${x_k}^{i}$.     
    The max-pooled vectors from all 21 blocks are vertically concatenated followed by $l_2$-normalization  to form the max-pooled image vector $x^{i}$. The   $x^{i}$ from all  images in Train1 and their corresponding image labels $l^{\,i}$  are used to train a linear SVM with weight $v$ and bias $b$ ~($\textbf{SW}=0$ in Fig.\ref{JointFramework}).         
          The classifier is used to predict the label $\hat{l}^{\,i}$ of  Train2 images.
 A misclassification $\hat{l}^{\,i}\neq l^{\,i}$ is an indication that the corresponding object model  has not been learned by the CRF comprehensively enough to include its appearance as in the misclassified image. Thus, the classifier selects those images with which the saliency model $w_n$ needs to be updated. Consequently, the corresponding object dictionary  $D_n$ is also updated, which, in turn, refines the global dictionary $D$ formed by concatenating the object dictionaries. 
  We use max-margin approach to identify the most violated constraints for the misclassified image and to update $D_n$ and $w_n$ accordingly (classifier-guided saliency update, $\textbf{SW}=0$ in Fig.\ref{JointFramework}). 
 
 After the above mentioned \textit{classifier-guided saliency update}, we improve the image classifier using saliency maps. The proposed  saliency-weighted max-pooling operation (\textbf{SW}=1 in Fig.\ref{JointFramework}) weights the max-pooled vectors ${x_k}^{i}$ of each block $k$ with its block saliency  $\tilde{s_k}^i$ . We infer the saliency maps from all training images Train1+Train2 and compute their saliency-weighted max-pooled $\tilde{x}^i$ vectors, which are used to train  a linear SVM $(\tilde{v}, \tilde{b})$. 
        This \textit{saliency-weighted image classifier} $(\tilde{v}, \tilde{b})$  is applied on a test image to refine the saliency map, and to classify it.
    At the end of the training stage, we get (i) a saliency model with updated CRF weight $w_n$  and dictionary $D_n$ and (ii) a saliency-weighted image classifier $(\tilde{v}, \tilde{b})$. 

 
  For test images, the saliency of each category,   $s_n^{i,j}$ is estimated by CRF inference using loopy belief propagation as in \cite{topdownCVPR2012}. 
  These saliency maps are refined using the posterior probability $\cal W$ estimated from the saliency-weighted  classifier  as explained in Sec.~\ref{sec:saliencyPostProcess}.  This \textit{refined saliency} ${RS}^{i,j}_n$   reduces the false detections in the saliency as shown in Fig.~\ref{fig:saliencyComparison}(h) and supported quantitatively in our experimental results. In summary, the proposed approach is able to simultaneously  identify and localize the object categories present in a test image.

\section{Category-aware sparse coding}
 
 The category-aware sparse coding  reuses the object dictionary $D_n$ and corresponding spare codes $z_n^{i,j}$   computed by the saliency component. The dictionary update in the saliency component improves the discriminative quality of the object dictionary $D_n$, and  a global dictionary $D$ formed by concatenating  updated  object dictionaries of all categories helps improve the image classification performance.  Moreover, this approach reduces the additional computations incurred by  forming a global dictionary $D$ using k-means clustering of thousands of patches from all categories followed by recomputation of sparse codes with respect to that dictionary.  
    Spatial pyramid max-pooling of the category-aware sparse codes (ScSPM) are used in our classification module. 
   \label{sec:Parallel-sparse coding}
     \begin{figure}[h]
     \begin{center}
      \includegraphics[scale=0.47, clip=true, trim=1.8cm 3.2cm 1.5cm 3.1cm]{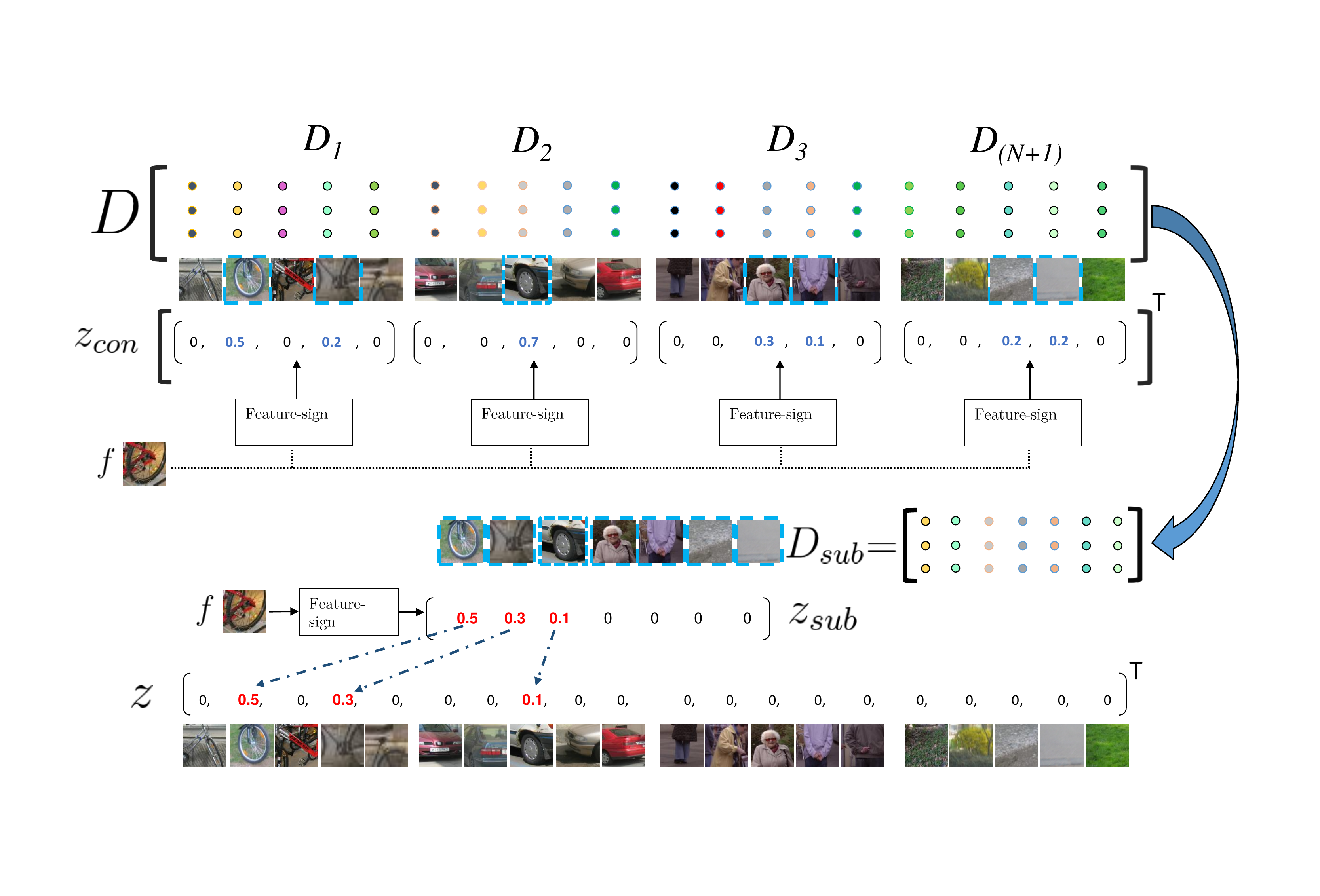}
     \end{center}
      \vspace*{-0.4cm}
     \caption{Illustration of category-aware sparse coding for classification (best viewed in color).}
      \label{dict_formation}
      \end{figure}
   Tight coupling between saliency modeling and image classification is obtained through the proposed category-aware sparse coding. It is to be noted that the following discussion pertains to computation of category-aware sparse code $z^{i,j}$ of a feature $f^{i,j}$. For simplicity, we drop the superscripts $i$ and $j$ in this section.
      Conventional sparse representation of a given feature ${f}$ aims to achieve the minimum  reconstruction error    while using sparse number of atoms from  $D$, i.e.,  
         \begin{equation}
         \label{eq:categSc}
          {z}=\underset{{z}}{\arg\min} \left \| {f}- D{z} \right \|_2  +\lambda\left \| {z}\right \|_1 .\\ 
          \end{equation}
    
 Since the objective of our image classifier is to improve saliency estimation, we introduce a category-aware  constraint to the feature coding  for classification, resulting in a novel category-aware sparse code $z$. i.e., the category-aware sparse code $z$ aims to achieve minimum reconstruction error while representing the feature $f$  with respect to each category dictionary $D_{n}$  as well as to the global dictionary $D$. 
  \subsection{Formulation}
   Let $D$ be the global dictionary formed by concatenating the individual object dictionaries, i.e. $D$ has a structure $[D_{1},~ D_{2},~\dots D_{(N+1)}]$, where $D_{n}$ represents each object dictionary and $(N+1)$ includes the $N$ object categories and the background class. The saliency model for an object category $n$ and its corresponding dictionary $D_{n}$ is learned as described in \cite{topdownCVPR2012}. The dictionary for the background class is formed by k-means clustering of background patches; thus, background features also have a sparse representation.  The size of $D$ is $k \times r_D$ where $r_D= (N\cdot r + r_{bg})$, $k$ is the dimension of the feature vector, $r$ and $r_{bg}$ are the number of atoms in the dictionary for each object class  and the background, respectively.
  

   The objective function for category-aware sparse coding is
 
 \begin{equation}
 \label{eq:Global_sc}
 z= \underset{z}{\arg\min} \left \| {f}- Dz \right \|_2  +\lambda_1 \left \| z\right \|_1+ \lambda_2\sum_{n=1}^{N}( \left \| {f}- DC_{n}z \right \|_2  +\lambda_3 \left \| C_{n}z\right \|_1) ,  
 \end{equation}  
     where the first two terms are the conventional  sparse coding of feature $f$ with $l_1 $ constraint  and the third term imposes our category-aware constraint.          
  
       $C_{n}$ is a selection matrix that selects the atoms of a particular object category $n$  from $D$. $C_{n}$ is derived from a zero matrix of size $r_D\times r_D$ by replacing its $m^{th}$  diagonal element with $1$, if  the $m^{th}$ atom in the dictionary $D$  belongs to the category $n$. For example if $D$ contains 6 atoms~($r_D=6$) with the third and fourth atoms of $D$ belonging to category $n$, then
       
           \begin{equation*}
           \centering
           C_{n}=
          \begin{bmatrix}
                      0& 0 & 0 & 0& 0& 0 \\ 
                        0& 0 & 0 & 0& 0& 0 \\ 
                       0& 0 & 1 & 0& 0& 0\\ 
                        0& 0 & 0 & 1& 0& 0\\ 
                                           0& 0 & 0 & 0& 0& 0 \\ 
                                            0& 0 & 0 & 0& 0& 0 \\ 
                      \end{bmatrix},
                     \end{equation*}
 and $C_{n}z$ selects the elements of $z$ that belongs to category $n$.  $\left \| {f}- DC_{n}z \right \|_2$ represents the reconstruction error for the sparse representation of feature ${f}$ with category dictionary $D_{n}$ (non-zero elements of $DC_{n}$= $D_{n}$) and the $l_1$-norm constraint $\left \| C_{n}z\right \|_1$ ensures that only few atoms of $D_{n}$ is used in this sparse representation. 
             
 \subsection{Approximate solution}
Since a closed-form solution of Eq.~(\ref{eq:Global_sc}) is not possible, we propose a computationally efficient approximate solution. Since the sparse codes  ${z}_n$  of each feature ${f}$ with respect to object dictionaries $D_n$ are already available from the saliency module, we develop a strategy to reuse these codes which  is computationally more efficient than conventional sparse coding of the feature  ${f}$  with global dictionary $D$. 
 
  The dictionary formation process and the category-aware sparse code vector ${z}$  are shown in Fig.~\ref{dict_formation}.   
The proposed solution minimizes Eq.~(\ref{eq:Global_sc}) in two steps. First, the  sparse codes ${z}_n,~\forall n\in \{1,...N\}$  of each feature ${f}$ with all category dictionaries $D_n ~\forall n\in \{1,...N\}$ are computed using feature-sign solver~\cite{featureSign}, which  reduces the third term in Eq.~(\ref{eq:Global_sc}), i.e 
 $\lambda_2\sum_{n=1}^{N}( \left \| {f}- DC_{n}z \right \|_2    +\lambda_3 \left \| C_{n}z\right \|_1) $. 
  
Let $z_n$ be the sparse code for feature ${f} $ evaluated with respect to object dictionary $D_n$.
 We form a vector $z_{con}$ by vertically concatenating the sparse codes of ${f} $ obtained for each $D_n$ as shown in Fig.~\ref{dict_formation}. The non-zero terms in $z_{con}$ point to atoms in $D$ that contribute to minimizing the reconstruction error of ${f} $ and hence, can effectively represent it. We pick these atoms from $D$ to form the classifier dictionary $D_{sub}$ and generate sparse code vector $z_{sub}$ for ${f} $ on $D_{sub}$ to minimize the overall objective function. 
 
 Since the number of non-zero terms in $z_{con}$ is small, the number of atoms in $D_{sub}$ is much lesser than in $D$ resulting in lesser computation for generating $z_{sub}$. 
 Since the number of atoms in $D_{sub}$ is different for every feature and the ScSPM classifier needs a dictionary of the same size as $D$, we need to represent $z_{sub}$ in a vector with the same length as $z_{con}$. 
 Since  $D_{sub}$ is formed by picking atoms from $D$, the elements of $z_{sub}$ can be placed in their respective locations of the category-aware sparse code $z$ having same length as the number of atoms in $D$, initialized with a zero vector. 
 Let a sparse code element $z_{sub_{p}}$ in the $p^{th}$ position in the code vector $z_{sub}$ correspond to atom $d_p$ in $D_{sub}$. If $d_p$ is the $m^{th}$ atom in $D$, then the code  $z_{sub_{p}}$ is placed in the $m^{th }$ position of category-aware sparse code $z$. The number of atoms in $D_{sub}$ for PASCAL VOC 2007 dataset~\cite{dataset_pascal-voc-2007} ranges from 300-400, which is much smaller compared to a typical classifier dictionary size for the same dataset, which ranges between 8000 and 12000~\cite{Vinay_IJCV}.
 
 \subsection{Computational Complexity}
We use the Feature Sign sparse solver (FS)~\cite{featureSign} whose time complexity to encode feature ${f}$ is 
$O(Lk)+O(LT_f)$
 when the dictionary size is $k \times L$ and $T_f$ is the sparsity  of the code (number of non-zero elements in the code). Using FS solver on the global dictionary $D$ (obtained by concatenating category dictionaries) 
  will result in a computational complexity of $O(r_Dk)+O(r_DT_f)$,  where $r_D=N\cdot r+r_{bg}$, which is very large for   datasets with a large number of classes (N). One possible approach to reduce the computation is to reduce $r_D$ by forming a smaller global dictionary by clustering of features from all categories (instead of concatenating category dictionaries). However, it may  result in loss of fine-grained information that helps in distinguishing similar classes. 

In the proposed framework, for every feature, FS solver is used in two stages - first, during saliency estimation, with respect to dictionaries $D_n$ and then with respect to sub-dictionary $D_{sub}$. 
Since there is no dependency between the first stage solvers,
a parallel implementation can effectively result in a time complexity of  $O(rk)+O(rc_{n})$ per feature for each object class  where $c_{n}$ is the sparsity of the  sparse code with $D_n$. Since $r_D$ is nearly $(N+1)$ times larger as compared to $r$, parallel implementation of proposed framework is $(N+1)$ times less complex as compared to the time complexity of conventional sparse coding on $D$.
 For the same sparse penalty $\lambda$, we have observed that $c_{n}$ in each category code is less than $T_f$, the sparsity with the global dictionary $D$. For sparse coding using $D_{sub}$, each feature requires an additional time complexity of $O(r_{sub}k)+O(r_{sub}s_{sub})$, where $s_{sub}$ is the sparsity of $z_{sub}$. Since  sparse codes are already available from the saliency estimation stage, the second round of sparse coding computations are the only additional computations required for sparse coding of classification, resulting in significant savings in computation.
     
\section{The image classification module}
 Classifiers are used for  saliency model training, saliency map refinement and for image classification. The proposed framework tightly couples these processes.  
%
%
%

 \subsection{Classifier to train saliency model (classifier feedback)}
 \label{sec:ClassifierTotrainSaliency}
The classifier uses 3-level spatial pyramid max-pooling~\cite{ScSPM_CVPR09} of category-aware sparse codes.
i.e, an image $i$ is divided into 21 blocks. Each block $k$ is represented by a single max-pooled vector $x_k^i$ of dimension $(N\cdot r+r_{bg})\times1$ formed by element-wise maximum of the category-aware sparse code vectors $z^{i,j}$ in that block. Each image  is represented with a max-pooled vector $x^i$ of dimension $21(N\cdot r+r_{bg})\times1$, formed by vertical concatenation of the max-pooled vectors from each block. 

 Let $\{\mathbf x^i, l^{\,i}\}\ , i \in \text{Train1}$ be the training data where $ x^i$ is the   $l_2$-normalized  vector from image $i$ and $l^{\,i} \in \{ 1, -1\}$ indicates the  presence or absence of the target object in that image $i$.   A one-vs-rest~(binary) linear SVM classifier  with SVM weight $v$ and bias $b$  is trained on half of the training images Train1 by minimizing following objective function ~\cite{liblinear}
  \begin{equation}
  \label{eq:svm_eq}
     \underset{v}{\arg\min} \|v\|^2 + C \hspace{-0.3cm}\displaystyle\sum_{i\, \in \, \text{Train1}}  \hspace{-0.3cm} max\,(0,1-l^{\,i} (v^\top x^i+ b)\;) ,  
    \end{equation} 
    where $C$ is the cost of constraints violation. 
 
 The max-pooled vector of each image $i$  from the other half of the training set, Train2, are used to validate the classifier using $f( x^i)=v^\top x^i+b$. Correct classification of image $i$ is indicated by $(f(x^i)\cdot l^{\,i})>0$. 
 The saliency model corresponding to the misclassified object is updated through  refinement of CRF weights and dictionary (classifier feedback). For example, if a bike image ($n$=bike) is misclassified by the bike classifier, the bike saliency model is updated using this image. 
  Let $Z_n^{\,i}= [z_{n}^{i,1},\ z_{n}^{i,2}\dots z_{n}^{i,t}]$ be the set of all sparse codes  of a misclassified image $i$ using  $D_n$.  and ${Y_n^{\,i}}=[y_{n}^{i,1},~y_{n}^{i,2},\dots y_{n}^{i,t}],~ y_{n}^{i,t}\in \{-1,1\} $ be the ground truth label for target $n$ presence in each patch.  Here $t$ is the total number of patches in the image. 
  If $\hat{Y}_n^{\;i}$ are the labels predicted  using category $n$ CRF model, 
 the loss function for the image is~\cite{topdownCVPR2012} 
   \begin{equation}
  \beta(w_n,D_n)=E(\hat{Y}_n^i,Z_n^i,w_n)-E(Y_n^i,Z_n^i,w_n),
      \end{equation}
where $E$ is the energy function  of  CRF built on a four connected graph conditioned on sparse codes $Z_n^i$ (for details  please refer to~\cite{topdownCVPR2012}). The ground truth energy $E(Y_n^i,Z_n^i,w_n)$ is less than any other energies $E(Y,Z_n^i,w_n)$ by a large margin $\Delta (Y,Y_n^i)$. i.e, 
 $E(Y_n^i,\,Z_n^i,\,w_n)\; \leq E(Y,\,Z_n^i,\,w_n)-\Delta (Y,\,Y_n^i) $ ~\cite{topdownCVPR2012}. Here $Y$ represents set of binary labels assigned to  patches in the the image $i$ and  $\Delta (Y,Y_n^i)=\sum_{j=1}^{t}\mathbb{I} (y^j,y_{n}^{i,j})$ indicates the margin function, where $\mathbb{I}$ is  an indicator function which is  $1$ when  $y^j\neq y_{n}^{i,j}$ and $0$ otherwise. 
 The most violated constraints are identified by solving 
   \begin{equation}
 \hat{Y}_n^{\;i}=\arg \min_Y E(Y,\,Z_n^i,\,w_n)-\Delta (Y,\,Y_n^i), 
       \end{equation}
 which is used to update the CRF weights $w$  for object $o$  as  \cite{topdownCVPR2012},
   \begin{equation}
   \label{eq:w_update}
   w_n=w_n- \rho_0 \frac{\partial \beta}{\partial w_n}, 
      \end{equation}

where $\rho_0=10^{\text{-}3}$ is the learning rate. 

Following ~\cite{topdownCVPR2012}, the updated CRF weight is used to update the  object dictionary $D_n$ for a given object $n$ (e.g bike) as

   \begin{equation}
   \label{eq:D_oUpdate}
   D_n=D_n+\rho_0 \frac{\partial \beta}{\partial D_n}.
      \end{equation}

\subsection{ Saliency-weighted classifier}

\begin{figure}[t]
    \begin{center}
                           \centering
                           \vspace{-0.3cm}
                        \mbox{ \centerline{ \includegraphics[width=0.9\textwidth, clip=true, trim=2.1cm 2.9cm 2.1cm 0cm] {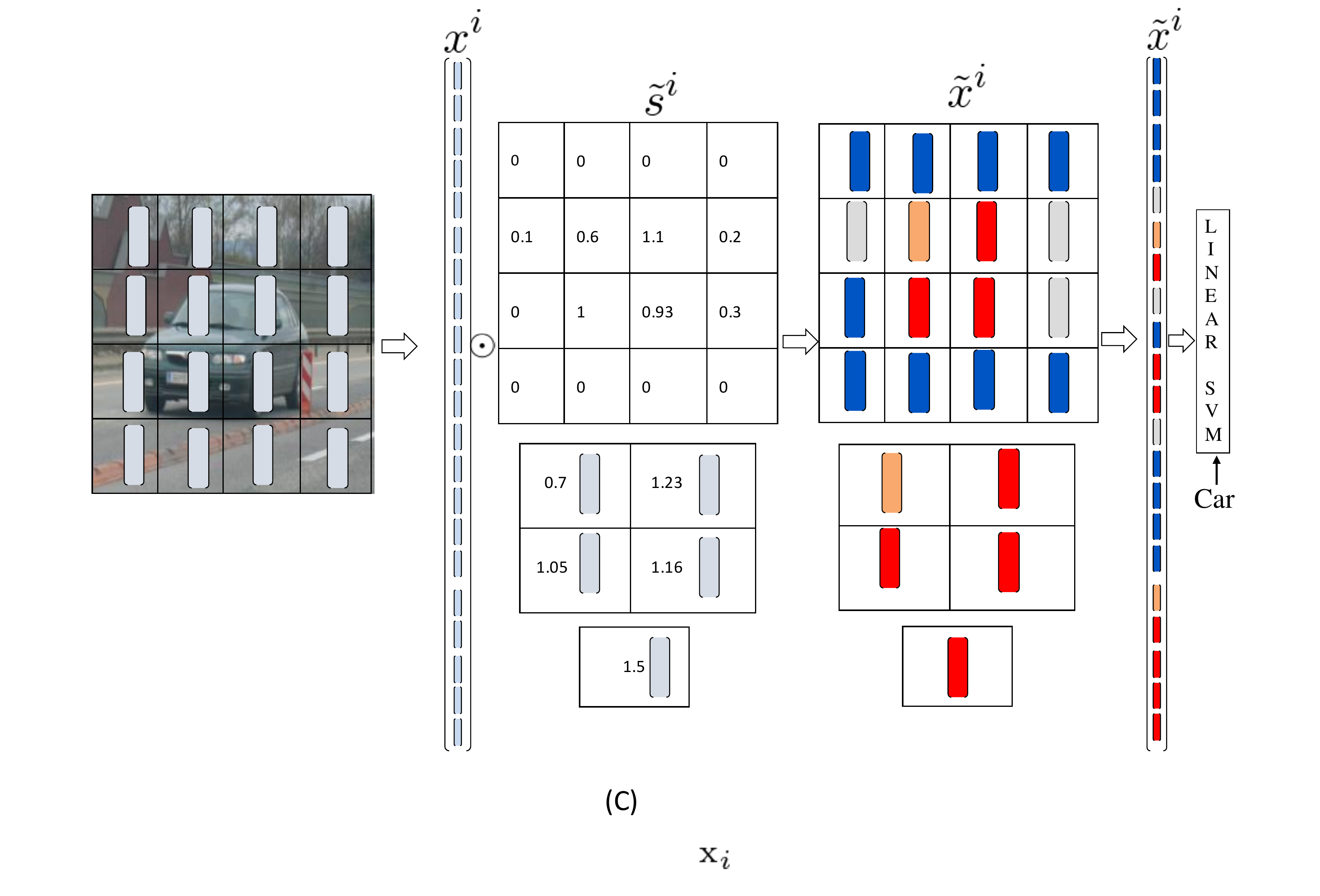}} 
                           }                      
                         \caption{ Illustration of saliency-weighted max-pooling (best viewed in color).}
                         \label{fig:saliencyWeightMaxpool}
                      \end{center}
     \end{figure}

Once the object dictionary refinement and saliency model learning is complete, we use saliency maps to improve classifier accuracy.  Fig.~\ref{fig:saliencyWeightMaxpool} illustrates the  saliency-weighted classifier  pipeline.
The max-pooled  vector at block $k$,  $x_k^i$  is weighted with the corresponding block-saliency value $\tilde{s}_k^i$ to form the saliency-weighted max-pooled vector $\tilde{ x_k}^i$ for that block. The saliency-weighted max-pooled vectors from all 21 blocks of the spatial pyramid are vertically concatenated to form the saliency-weighted image vector $\tilde{ x}^i$. 
An image classifier is learned to indicate the presence or absence of target object ($l^{\,i} \in \{ 1, -1\}$) using  $\tilde{x}^i$  from all the training images, i.e (Train1+Train2). 
\label{sec:saliencyWeightedClassifier}

\subsubsection{Block saliency computation}
  \begin{figure}[t]
    \begin{center}
                          \centering
                           \vspace{-0.3cm}
                            \mbox{ \centerline{  \includegraphics[width=1\textwidth, clip=true, trim=0.2cm 16.7cm 3.4cm 0.2cm]{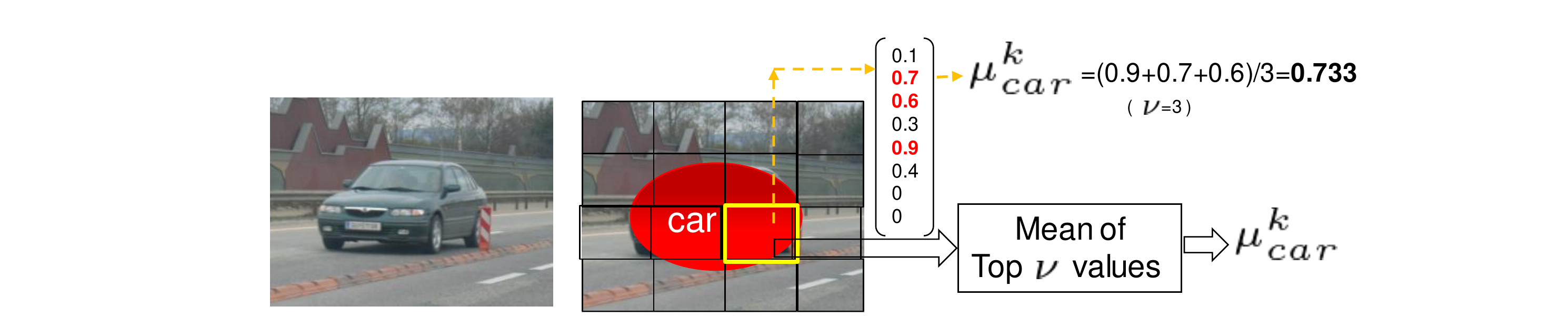}}} \\
                                  (a)
                             \mbox{ \centerline{  \includegraphics[width=1\textwidth, clip=true, trim=0.2cm 4.5cm 3.0cm 8.2cm]{saliencyWeighting1C.pdf}}} \\
                               \hspace{0.2cm} $  \mu_{car}$ \hspace{2.5cm} $\mu_{bike}$ \hspace{2.0cm} $\mu_{person}$ \hspace{2.9cm} $\tilde{S}^i$\\

                             (b)
                     
                         \caption{ Illustration of block saliency computation for blocks of spatial pyramid  (best viewed in color).}
                         \label{fig:BlocksaliencyWeight}
                      \end{center}
     \end{figure} 
   The saliency model of a particular class highlights those patches that are likely to contain object parts belonging to that class. High saliency values indicate either object regions in a positive image or possible  false positive patches in a negative training image. 
Weighting corresponding max-pooled sparse codes with this high value and training the SVM with its corresponding image label will help  the image  classifier to reduce false detections and  improve its performance  against background clutter. To this end, for each image, our objective is to determine a saliency weight for the max-pooled vector in each of the 21 spatial blocks of the pyramid, using the $N$ saliency maps computed for the $N$ categories.

The first step involves finding the saliency value for each category in each block of the spatial pyramid. Choosing the maximum saliency  for each block would lead to a poor representation caused by outliers in the saliency map. Using the mean saliency of an object within a block may result in loss of saliency information at higher levels of the pyramid. This is attributed to the presence of many background pixels with low saliency values  that are present in the larger image areas covered by these levels.
Hence, we use the mean, $\mu$, of the top $\nu$, ($\nu=2$) saliency values of an object category  in a block as shown in Fig.~\ref{fig:BlocksaliencyWeight}(a).

The block saliency $\tilde{s_k}^i$, which is the weight of the max-pooled sparse code vector for the $k^{th}$ block is obtained as $\tilde{s_k}^i=\sum_{n=1}^N \mu_n^{k}$  (Fig.~\ref{fig:BlocksaliencyWeight}(b)).
 Since we use mean of top saliency values in a block corresponding to all object models, background patches are suppressed compared to the object patches.
\subsubsection{Saliency-weighted  max-pooling}
 Our weighting criteria is simple and computationally efficient enabling its use in larger datasets having multiple objects in an image.
  Every element of the max-pooled vector from a block $k$ is multiplied with its corresponding block saliency $\tilde{s}_k^i$ to form the saliency-weighted max-pooled vector $\tilde{x}^i_k$  for that block.
   i.e 
    \begin{equation}
    \label{eq:SalWeight}
 \tilde{ x}^i_k =\tilde{s}_k^i \cdot  {x^i}_k ,\hspace*{0.4cm}  \forall k \in \{1,2,\dots 21\} 
  \end{equation}

%

   The  $l_2$-normalized saliency-weighted max-pooled vectors  $\tilde{x}^i_k$ from entire training set~(Train1+Train2) are used to learn a one-vs-rest~(binary) linear SVM classifier $(\tilde{v},\tilde{b})$ for image classification as in Eq.~(\ref{eq:svm_eq})~\cite{liblinear}.
  

 \begin{equation}
     \underset{\tilde{v}}{\arg\min} \|\tilde{v}\|^2 + C \hspace{-0.7cm} \displaystyle\sum_{i \; \in (\text{Train1+Train2})}  \hspace{-0.7cm} max \; (0,\;1-l^{\,i} \text{(}\tilde{v}^\top \tilde{x}^i+ \tilde{b})\; \text{)} ,
    \end{equation}

\section{Saliency inference and refinement}
\label{sec:saliencyPostProcess}
 The final saliency map generated by the proposed method is a weighted version of the one proposed in \cite{topdownCVPR2012} where the weights are based on the output of the saliency-weighted classifier explained in Sec.~\ref{sec:saliencyWeightedClassifier}. 
  In \cite{topdownCVPR2012}, a four connected graph having Markov property is formed on image patches based on their spatial adjacency.
 The probability of label $y_{n}^{i,j} \in \{-1,1\}$ indicating the absence  or  presence of  object $n$ respectively at a patch $j$ is computed from its neighbours using marginal probability \cite{topdownCVPR2012}
 \begin{equation} 
  P(y_{n}^{i,j}\;|\;z_{n}^{i,j},\;w_n)=\sum_{y_n^{i,{\cal N}(j)}}p(y_n^{i,j},\;y_n^{i,{\cal N}(j)}\;|\;z_{n}^{i,j},\;w_n)\,,
    \end{equation}
 where $z_{n}^{i,j}$ is the sparse code of a patch $j$, $w_n$ is the CRF weight vector and ${\cal N}(j)$ is the neighbourhood of node $j$ with label $y_n^{i,{\cal N}(j)}$.

 Saliency of a patch $j$ is given by 
\begin{equation}
 \label{eq:sal_level1} 
  {s_n}^{i,j}= P({y_n}^{i,j}=1|z_{n}^{i,j},w_n),
  \end{equation}
which is inferred using loopy belief propagation~\cite{topdownCVPR2012}. The  neighborhood and visual priors  of SIFT features are used in this saliency computation. 
The refined saliency ${RS_n^{i,j}}$ for patch $j$ is 
\begin{equation}
 \label{eq:sal_refine}
RS_n^{i,j}={s_n}^{i,j} \cdot{\cal W},
\end{equation}
   where ${\cal W}$ is derived from SVM confidence $(\tilde{v}^\top \tilde{ x}^i+ \tilde{b}) $ of the saliency-weighted classifier. Since the weight is the same for all patches, it is equivalent to multiplying the saliency map of the image with the weight. The spatial prior is used by the  saliency-weighted classifier  for the computation of ${\cal W}$ through its spatial pyramid max-pooling operation. ${\cal W}$ is assigned  a high value if the location of the object of interest in a test image  matches its  spatial prior (object  location in the training data).  Objects appearing at spatial locations which are significantly different from their training data  are assigned  lower ${\cal W}$  values. 

    The training images of a dataset indicate the maximum number of object categories in an image. If there is only 1 object category per image and  the class with highest SVM confidence matches  the model being tested, $\mathcal{W}=1$ and the generated map becomes the final saliency map. 
         If there are multiple object categories per image, 
        we evaluate the classifier confidence  for each of the object categories within a test image and sort them in descending order. If the confidence of an desired object category is ranked below the maximum number of objects per image in the training set,
        it is highly unlikely that object is present in the  image and hence the refined saliency map will have no salient patches in that test image. If the confidence of the object category is ranked within the highest number number of objects per image, a rescaled classifier confidence serves as  ${\cal W}$.
    In Graz-02 dataset~\cite{Graz_02_dataset}, which has only one object class per image, the saliency map corresponding to the object class predicted by the saliency-weighted classifier is weighted by 1; the saliency maps of the other categories are weighted by 0. While this strategy of weighting the saliency map might appear to be totally dependent on the classifier performance, it must be noted that the classifier uses saliency-weighted pooling so that an accurate saliency map will contribute to lowering the error in classification. 
    In PASCAL VOC-07 which has 20 classes, there is no training image with more than 5 object classes. If the classifier confidence for a particular class is ranked more than 5, ${\cal W}$=0. Since most of the training images have 1 or 2 object classes, ${\cal W}$=1 if the classifier confidence is ranked first or second. For the remaining positions, ${\cal W}$ is taken as the linearly scaled classifier confidence.

 \section{Experimental Results}
\label{sec:experimental}
 We evaluated the performance of our joint framework on Graz-02 and PASCAL VOC 2007 datasets. 
     SIFT features are extracted from $64\times64$ patches with a grid spacing of 16 pixels as in \cite{topdownCVPR2012}. We choose $\lambda$ in the sparse code formulation $z_{n}=\arg \min_{z_n} \ \|{f}-D_n z_n\|_{2}+\lambda\ \|{z_n}\|_{{1}}$ to be 0.15 for both classifier and saliency modules. We evaluate top-down saliency using precision rates at equal error rate (EER). The saliency maps are thresholded at 100 levels between 0 and 1 and a precision recall graph is drawn. The EER refers to the point at which the precision is equal to the recall. 
 \subsection{Training and testing image selection }    
 
\subsubsection*{Graz-02 dataset}
  Graz-02 has 3 object categories- bike, car and person, each having 300 images with pixel-level object annotations and an additional 65, 120 and 11 images without object annotations in each category respectively. Apart from this 380 background images are also present.\\
We conduct the experiments on different sets of training and test image combinations. For comparison of the proposed top-down saliency with recent top-down saliency approaches, we only consider  object annotated images  for training and testing. Following~\cite{topdownCVPR2012,topdownBMVC2014}, odd numbered images are used for training the proposed saliency model and even numbered images for testing.\\
Secondly, to compare classifier performance with related image classifiers, the same training and test images selection procedure in Bilen \emph{et al.}~\cite{Vinay_IJCV} is used. All~(1096) object images  are used for classification. For each category, 150  training images are selected at random, and remaining are used for testing. The average results of 10 such experiments are reported.
\subsubsection*{PASCAL VOC-07 dataset}
 
 PASCAL VOC 2007 is a challenging dataset consisting of 20 different categories 
      with some images having objects from multiple object categories.
       Similar to \cite{topdownCVPR2012}, we evaluated the performance of our saliency model on 210 segmentation annotated test images. Since there are only 422 segmentation  images to train 20 categories, we use additional images from the  object detection challenge as in \cite{topdownCVPR2012} (nearly 150 images per category). 
     For comparison with related classifier approaches, the training, validation, and test image combinations in the \emph{classification challenge} of the dataset is used.

 \subsection{Top-down saliency }
\subsubsection{{Graz-02 dataset}}
 For each object category, an object dictionary with 512 atoms and  corresponding CRF parameters  are learned for 10 iterations. 
   From $3\times512=1536$ object atoms and 512 background atoms, a global dictionary of 2048 atoms is formed by concatenation. The ground truth label of a patch is 1 if at least 25\% of its pixel belongs to object of interest, and -1 otherwise.   In \cite{topdownCVPR2012}, training for each category is done using 150 positive images and 150 background images (column T1 in Table~\ref{table:Trainingset}). To improve performance against the false positive detection, training of proposed framework uses negative images from other categories as well~(T2 in table.~\ref{table:Trainingset}). Since T2 has 150 positive images and 450 negative images, the training set to re-train~\cite{topdownCVPR2012} is balanced by randomly selecting 150 negative images from 450 negative images available in T2. 
   
       Since the proposed framework requires training and validation cycles,  T2 is divided into T2a~(training set) and T2b~(validation set). 
     T2a contains 70 positive images for each category and 80 negative images, 30 of which are from background and 25 each from the other 2 categories. Since 70 positive images from each category are used to form T2a, the remaining 80 images per category form 320 images in T2b. Misclassified images from T2b are used to update the top-down saliency model. 
   
  Table~\ref{table:Graz_saliency}(a) compares the patch-level precision rates at EER of the proposed saliency detection with \cite{topdownCVPR2012} on the Graz-02 dataset.
  In \cite{topdownCVPR2012}, the authors tested each saliency model on  object annotated  images from its respective category and background.
   An ideal top-down saliency estimator should be able to distinguish the object from background clutter as well as from other objects. 
   So, we tested  \cite{topdownCVPR2012} on  images from all categories in the dataset,
   i.e, we use their model trained on T1 and evaluated on all 600 images. The average EER of 54.9\% is less than 73.7\% reported in \cite{topdownCVPR2012}. Since our training mechanism involves negative images from other categories in addition to the background category, we used T2 to train~\cite{topdownCVPR2012}. The increase of about 5\% in EER illustrates the utility of negative images for training.

  \begin{table}[t]
       \centering
   \caption{Graz-02: Saliency training  sets used in our experiments.}
                 \label{table:Trainingset}
                  \resizebox{0.75\textwidth}{!}{
  \begin{tabular}{@{}lllll@{}}
  \toprule
  Image set                                     & T1  & T2    & T2a   & T2b   \\ \midrule
  Number of positive training images          & 150 & 150   & 70   & 80   \\
   Number of background images                & 150 & 150   & 30   & 80   \\
  Number of negative images  except  background & 0   & 2x150 & 2x25 & 2x80 \\  \midrule
  Total number of training images per category  & 300 & 600   & 150  & 320  \\ \bottomrule
  \end{tabular}
  }
  \end{table}

\begin{table}[t]
             \centering
            \caption{Precision rates at EER (\%) of proposed method against other top-down saliency approaches on all~(600) test images of Graz-02 dataset.
           }
            (a) \textbf{{Patch-level}\\}
             \label{table:Graz_saliency} 
         \resizebox{0.9\textwidth}{!}{  
\begin{tabular}{@{}cccccc@{}}
\toprule
Algorithm        & \begin{tabular}[c]{@{}l@{}}Yang and  \\ Yang~\cite{topdownCVPR2012}\end{tabular} & \begin{tabular}[c]{@{}l@{}}Yang and \\ Yang~\cite{topdownCVPR2012}\end{tabular} & \begin{tabular}[c]{@{}l@{}}Proposed \\  method\end{tabular} & \begin{tabular}[c]{@{}l@{}}Proposed\\   method \end{tabular} & \begin{tabular}[c]{@{}l@{}}Proposed method on a \\ classifier of 100\% accuracy \end{tabular} \\ \midrule
Training Set     & T1                                                                & T2                                                              & T2a                                                        & T2           & T2                                              \\
Number of  trg. iter. & 20                                                                & 20                                                              & 10                                                         & 10+2                             & 10+2                          \\ \midrule
Bike           & 62.5                                                              & 69.4                                                            & \textbf{75.6 }                                                      & \textbf{75.6}                              & 79                         \\
Car              & 53.6                                                              & 53.2                                                            & 53.8                                                       & \textbf{58.3}                           & 65.8                            \\
Person           & 48.6                                                              & 57.2                                                            & 62.8                                                       & \textbf{64.5 }                            & 71                          \\\midrule
Mean             & 54.9                                                              & 59.93                                                           & 64.06                                                      & \textbf{66.13}                     &71.9                                 \\ \bottomrule
\end{tabular}
}
\\
\vspace{0.2cm}

(b) \textbf{\small{Pixel-level}}\\
%
   \resizebox{0.6\textwidth}{!}{
\begin{tabular}{@{}llll@{}}
\toprule
Algorithm             & \begin{tabular}[c]{@{}l@{}}Yang and \\ Yang  \cite{topdownCVPR2012}\end{tabular} & \multicolumn{1}{c}{\begin{tabular}[c]{@{}c@{}}Kocak \textit{et al.}\\ \cite{topdownBMVC2014}\end{tabular}} & \begin{tabular}[c]{@{}l@{}}Proposed\\  method\end{tabular} \\ \midrule
Training Set          & T2                                                                & T2                                                                                & T2                                                         \\
Number of  trg. iter. & 20                                                                & 20                                                                                & 10+2                                                       \\ \midrule
Bike               & 59.43                                                             & 59.92                                                                             & \textbf{64.4}                                              \\
Car                   & 47.36                                                             & 45.18                                                                             & \textbf{50.9}                                              \\
Person                & 49.82                                                             & 51.52                                                                             & \textbf{56.4}                                              \\\midrule
Mean                  & 52.2                                                              & 52.21                                                                             & \textbf{57.23}                                             \\ \bottomrule
\end{tabular}
}
\end{table}


\begin{table}
 \centering
       \caption{Precision rates at EER (\%) of proposed method against other localization approaches on 150 test images of Graz-02 dataset.}
         \label{table:Graz_salPix_150}
  \resizebox{0.7\textwidth}{!}{
 \begin{tabular}{lccc}
 \hline
 Algorithm & Fulkerson \textit{et al.} \cite{fulkerson2008localizing} & \begin{tabular}[c]{@{}c@{}}Marszalek and \\ Schmid~\cite{shapeMaskIJCV2012}\end{tabular} & \multicolumn{1}{l}{Proposed method} \\ \hline
 Bike & 66.4 & 61.8 & \textbf{67.3} \\
 Car & 54.7 & 53.8 & \textbf{59.8} \\
 Person & 47.1 & 44.1 & \textbf{57.1} \\\midrule
 Mean & 56.07 & 53.23 & \textbf{61.4 }\\ \hline
 \end{tabular}
 }
 \end{table}

   \begin{figure*}[ht]
   \begin{center}
    \centering
    \includegraphics[scale=0.49, clip=true, trim=2.5cm 2.5cm 2.5cm 0cm]{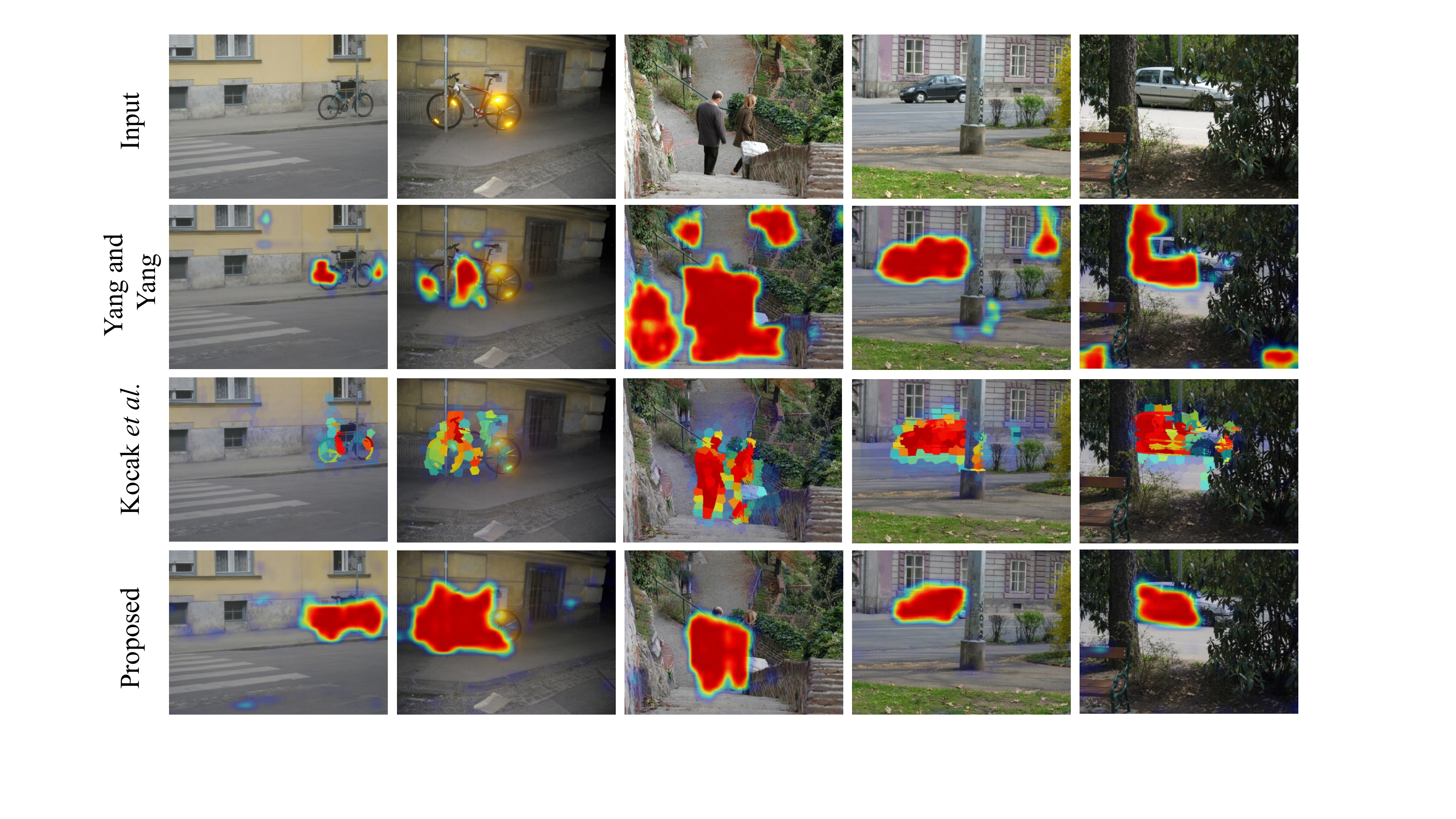}
 %
   \caption{Qualitiative comparison of saliency maps of Yang and Yang~\cite{topdownCVPR2012} and Kocak \textit{et al.}\cite{topdownBMVC2014} with the proposed method.}
    \label{fig:saliency_qualitative}
     \end{center}
    \end{figure*}
 \begin{figure}[h!]
     \begin{center}
                            \centering
                             \mbox{ \centerline{  \includegraphics[width=0.8\textwidth, clip=true, trim=0.2cm 14.7cm 14.8cm 0.2cm]{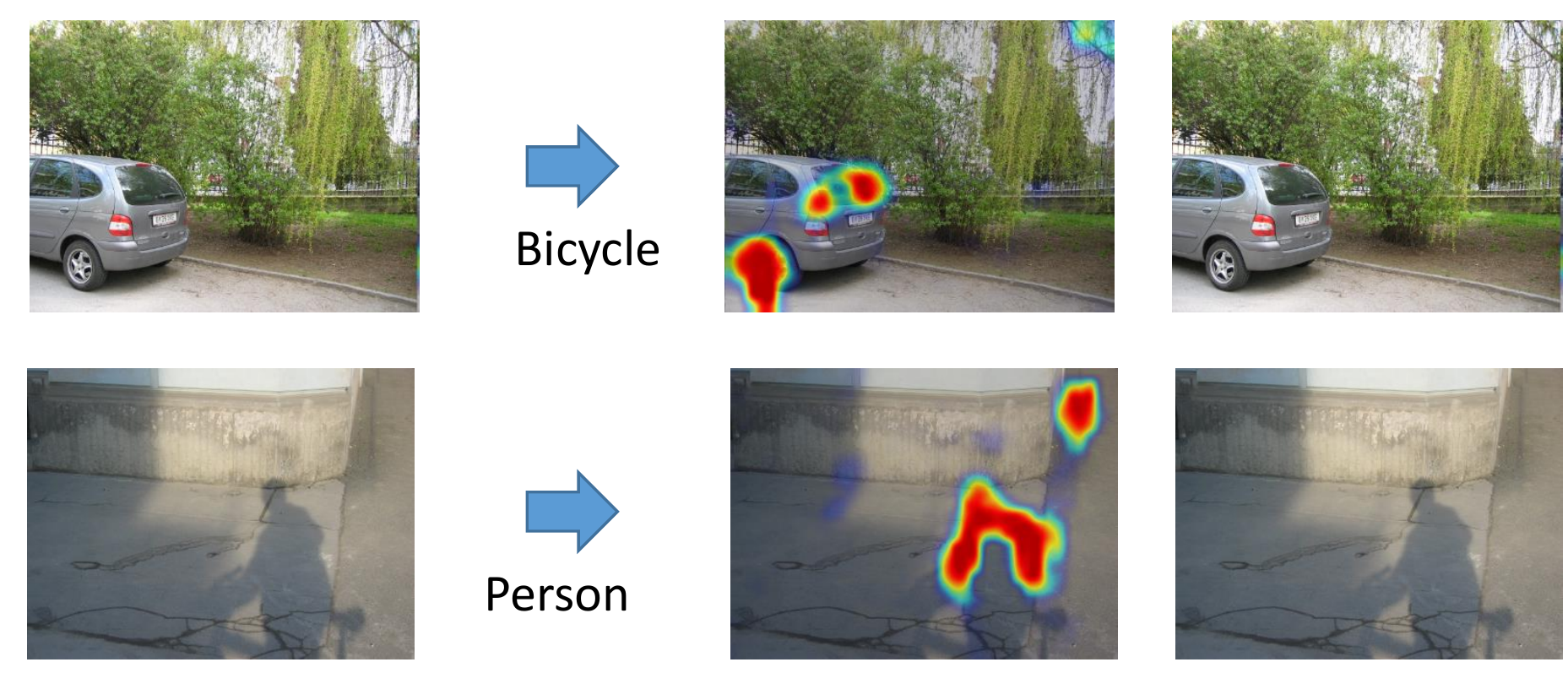}}}
                              
                             \hspace{0.3cm} (a) \hspace{4.3cm} (b)   \hspace{2.5cm} (c)
                          \caption{Removal of false detections on negative images by saliency refinement.(a) Input image, (b) false detections of bicycle model (row 1) and person model(row 2) before saliency refinement, (c) saliency-refined image with no false detections}
                          \label{fig:ReftWithWithoutSucc}
                       \end{center}
      \end{figure}
       \textbf{\textit{Effect of saliency refinement.}}  Our saliency modeling process has 2 stages - in the first 10 iterations, the models are learned using T2a and in the second stage the classifier feedback improves the models through misclassified images. In this experiment, we consider only the first stage and demonstrate the utility of refining the saliency map in table~\ref{table:Graz_saliency}(a)~(column~3). The saliency models trained using T2a for 10 iterations resulted in a mean precision rate at EER of  54.1\% without saliency refinement and without classifier feedback. 
        Using saliency model and dictionaries obtained at this $10^{th}$ iteration of saliency modeling, we train a saliency-weighted image classifier~(Sec.~\ref{sec:saliencyWeightedClassifier}) that uses multi-class   SVM~\cite{cramer_singer_svm} on the entire training images. On test images, the saliency maps are weighted using classifier confidence output for that image as  described in Sec.~\ref{sec:saliencyPostProcess}. There is a gain of 10\%  in precision rates at EER due to the saliency refinement and the resulting model outperforms \cite{topdownCVPR2012} with a  5\% gain in precision rates at EER. 
        Moreover, we use only 150 training images and 10 iterations when compared to 300 training images and 20 iterations in \cite{topdownCVPR2012} for saliency modeling. The selection of relevant atoms from the dictionary that can represent the feature for its sparse representation and a saliency-weighted classifier trained on the category-aware sparse codes jointly contribute to improve the performance. 
 
 The bicycle model causes few false detections  on  the \textit{car} image  of Fig.~\ref{fig:ReftWithWithoutSucc}, due to the similarity in the structure. For example, the round shaped tyre patches in \textit{car} is detected due to the similarity with the   bicycle tyre. Since the image classifier uses  high level information in the image, it could better predict that bicycle is absent in the given image and it removed the false detections as shown in Fig.~\ref{fig:ReftWithWithoutSucc} (c).
  Similarly, the shadow of human in the \textit{background} image introduced false detections while inferring a person model.  Again, classifier-based saliency refinement removed those false detections.
 
   \begin{figure}[h!]
       \begin{center}
                              \centering
                               \mbox{ \centerline{  \includegraphics[width=1\textwidth, clip=true, trim=0.0cm 13.9cm 8.9cm 0.0cm]{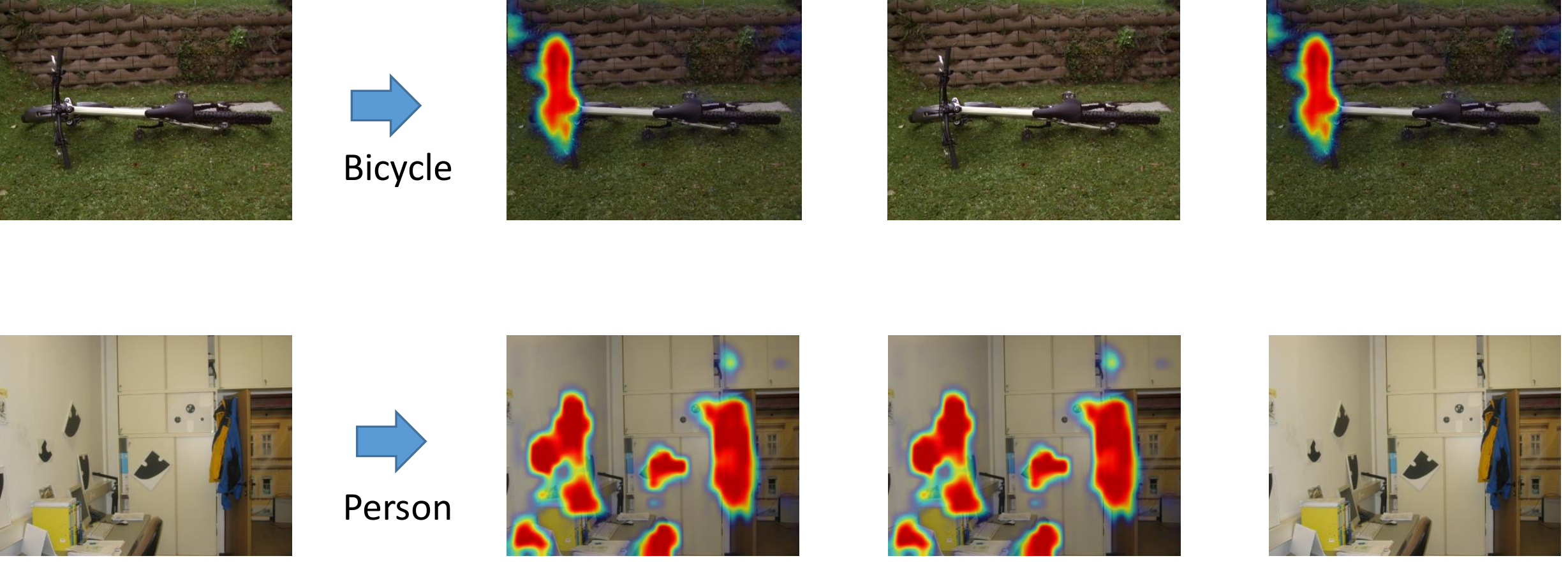}}}\\
                                   \hspace{0.1cm} (a) \hspace{3.7cm} (b)   \hspace{2.6cm} (c) \hspace{2.7cm} (d)  \\
                            \caption{ Failure cases of saliency refinement. (a) Input image, (b) true detections of bicycle model (top row) and  false detections of person model (bottom row) before saliency refinement. The misclassifications of the classifier leads to errors in the final saliency map (c)  with false negative in the top row and false positive in the bottom row for the bike and person models respectively. (d) An image classifier with 100\% accuracy could avoid both of these errors}

                            \label{fig:ReftWithWithoutFail}
                         \end{center}
        \end{figure}
 
 
 \textit{Failure cases of the saliency refinement.}
Saliency refinement largely depends on the accuracy of the image classifier. It may introduce  two types of errors in the saliency map, (i) false negative, when the image classifier wrongly predicts the absence ($\cal{W}$$=0$)  of the object in a  positive image Fig.\ref{fig:ReftWithWithoutFail} (top row)  (ii)  false positive, when the image classifier wrongly predicts the object presence in a negative image  Fig.\ref{fig:ReftWithWithoutFail} (bottom row). Due to viewpoint changes, the  image classifier fails to predict bicycle presence in the first case and in the second image, shirt hanging on the door matches with the position and features of person class. However, there are very few  such errors in our model.

\textit{Performance on  100\% accurate image classifier.}
    A 100\% accurate image classifier will assign $W=1$ for positive test images and   $W=0$ for negative test images. To evaluate the performance under 100\% accuracy, we manually assigned these values to $W$, based on the ground-truth of the test image. 
  Such an image classifier with 100\% accuracy can remove both these errors and produce a better saliency map as shown in Fig.\ref{fig:ReftWithWithoutFail} (c). Such a classifier improves our saliency accuracy on 600 test images to 71.9\% (last column in table~\ref{table:Graz_saliency}(a)), which is closer to the 73.7\% reported in \cite{topdownCVPR2012} for 300 test images. The slight degradation in accuracy is due to the use of less number of training cycles in our model, to save training time. To achieve a faster training of the model, we used only 10 initial iterations of the saliency model training compared to 20 in \cite{topdownCVPR2012}. Moreover, in these 10 iterations, we used only 150 training images in Train1 compared to 300 training images of \cite{topdownCVPR2012}. With these two changes, our initial training time reduced to 30\% of the time in \cite{topdownCVPR2012} (10 iterations using 150 images vs 20 iterations using 300 images).
  
    \textbf{\textit{Effect of classifier feedback.}} In this experiment, we study the second stage of saliency modeling. The dictionaries obtained at the end of the first stage of saliency modeling are used to train the multi-class classifier using T2a training set and validated using T2b. Misclassified images from T2b are used to train the corresponding saliency model. Another iteration of this feedback is carried out by interchanging T2a and T2b. There is an improvement of  2\% in precision~(column 4 of table    \ref{table:Graz_saliency}(a) ) due to classifier feedback, which is attributed to the car and person classes. The bike class is, by far, the easiest to model among the three as seen in \cite{topdownCVPR2012} also, and hence the initial saliency models are able to capture the variability in most of the images. However, using the failed images to provide feedback from the classifier was not sufficient to improve the bike models. Compared to \cite{topdownCVPR2012}, there is an improvement of about 7\% with only 10 iterations for saliency modeling and 2 iterations for classifier feedback. The results are supported qualitatively too, as shown by the saliency maps in Fig.~\ref{fig:saliency_qualitative}. 
   
  \begin{figure}[h!]
        \begin{center}
                               \centering
                                \mbox{ \centerline{  \includegraphics[width=0.8\textwidth, clip=true, trim=0.1cm 10.2cm 15.6cm 0.2cm]{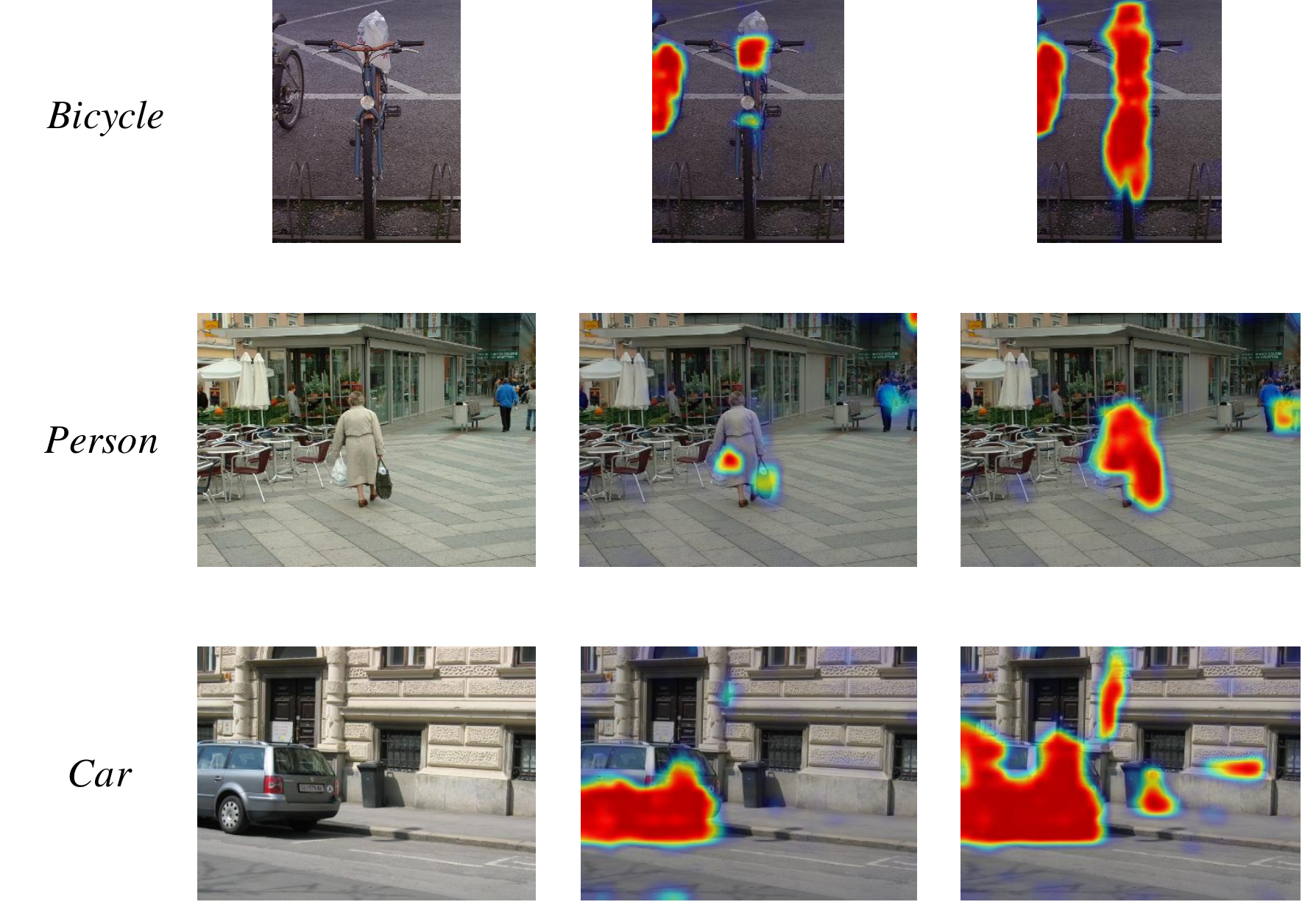}}} \\
                                  \hspace{1.8cm} (a) \hspace{2.4cm} (b)   \hspace{2.5cm} (c)
                             \caption{Effectiveness of classifier feedback. (a) Input image, (b) saliency map of a model before classifier feedback and (c) saliency map of a model trained with classifier feedback. The \textit{bicycle} and  \textit{person} images (c) show improvement in the saliency map due to feedback, while the \textit{car} image (c) shows a failure case, where the classifier feedback introduced few false positives. }
                             \label{fig:ClassFbkWithandWitout}
                         \end{center}
         \end{figure} 
   
In Fig.~\ref{fig:ClassFbkWithandWitout}, the saliency maps of \textit{bicycle} and \textit{person} images with distinct pose/features are improved by the classifier feedback, while on \textit{car} image, even though the refinement improved the true positive detection of the saliency map, it also introduced some false detections along straight edges of the background clutter.
Inspite of few exceptional cases, the classifier feedback helps to improve the saliency map especially on the images which has a rare pose or view point as shown in  \ref{fig:ClassFbkWithandWitout} (c) (\textit{bicycle} , \textit{person}). This is justified quantitatively too in Table~\ref{table:Graz_saliency}.

\textbf{\textit{ Pixel-level result comparison.}} For comparison with a recent top-down saliency approach~\cite{topdownBMVC2014}, we used their publicly available code and re-trained their models using T2 and evaluated on entire 600 test images. Pixel-level precision rates at EER~(table~\ref{table:Graz_saliency}(b) ) show that the proposed patch-based approach achieves state-of-the-art result at pixel-level as well. Note that pixel-level saliency maps are generated from patch-level saliency maps as described in \cite{topdownCVPR2012}. Despite using computationally intensive characteristics like 'objectness', and extracting color-based features from every superpixel, the performance of~\cite{topdownBMVC2014} drops when evaluated on the entire 600 test images, instead of evaluating on 150 same category and 150  background test images. This can be attributed to the inability to discriminate between object categories.
 
\textbf{\textit{ Comparison with other approaches.}}\
Our top-down saliency results are also compared with object localization/segmentation approaches that use precision rates at EER as evaluation metric. 
   The proposed saliency models are evaluated on the test image set-up of shape mask~\cite{shapeMaskIJCV2012}, i.e, each model is evaluated only on 150 test images from their
   respective category.
    Table~\ref{table:Graz_salPix_150} shows
  the effectiveness of proposed saliency model  by outperforming \cite{fulkerson2008localizing} and \cite{shapeMaskIJCV2012} in all the three categories. It is to be noted that  \cite{shapeMaskIJCV2012} needs an additional level of supervision by manually labeling training images as \emph{truncated} or \emph{difficult}.
Object class segmentation~\cite{fulkerson2009class} extends ~\cite{fulkerson2008localizing} using superpixels as the basic unit for computation and the segmentation results are refined using a CRF operating on the superpixel graph. By maintaining identical parameters as our approach~(without aggregating the histograms of a superpixel with its neighboring superpixels), ~\cite{fulkerson2009class} achieves a mean precision at EER of 54.56\% which is lower than the proposed method operating on regular rectangular patches~(61.4\%).

  \subsubsection{{PASCAL VOC 2007 dataset}}
  
We use all the available positive training (P1) and validation (P2) images to train the initial saliency models for 10 iterations.
       Since multiple classes are present in some images, we train  one-vs-rest binary classifiers for each object using P1 images 
       and validate on  P2. Set of positive images with lower confidence are used  to train respective saliency model incrementally.
  \begin{figure*}[t]
    \centering
  \includegraphics[width=14cm, clip=true, trim=7cm 9.75cm 8cm 5cm]{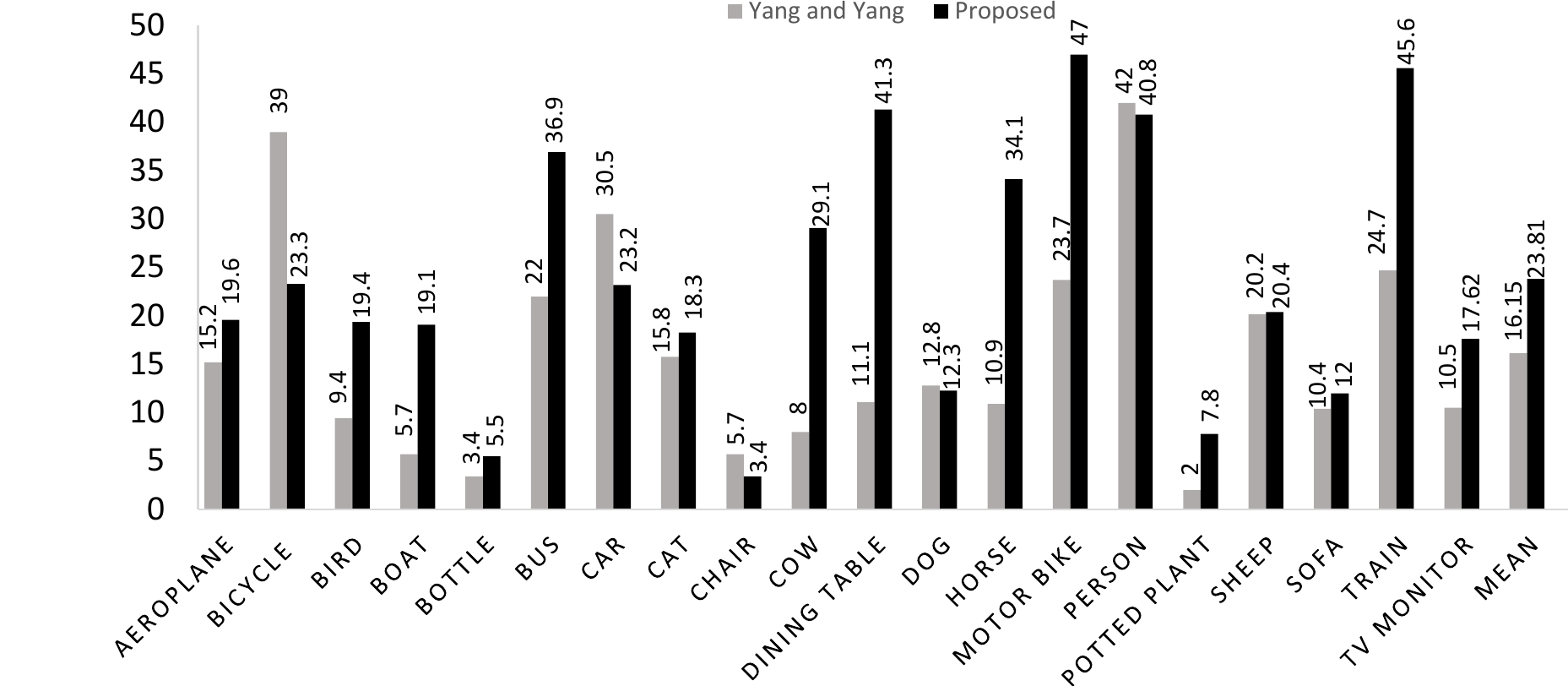}
  \hspace{2cm}  \caption{Patch-level precision rates~(\%) at EER on PASCAL VOC-07 compared to Yang and Yang~\cite{topdownCVPR2012}.}
  \label{fig:VOCResult}
  \end{figure*} 
  
    Fig.~\ref{fig:VOCResult} shows results for patch-level saliency estimation in which we achieve state-of-the-art performance in 15 out of 20 classes. Averaging over the entire 20 classes, we achieve a mean precision rate at EER~(\%) of 23.81  compared to 16.15 of~\cite{topdownCVPR2012}, which is an improvement of 47\%. 
    We maintained the same dictionary size (512) and number of CRF weights as in \cite{topdownCVPR2012}. The proposed saliency model trained for less number of cycles performs much better than \cite{topdownCVPR2012} trained for 20 iterations. This is attributed to \cite{topdownCVPR2012} failing in images having large dominant objects like bus or horse, as local patches of these contain limited relevant information. The additional classifier-guided dictionary training in the proposed model is able to capture this information leading to a significantly higher precison rate at EER for such classes. \\
             \begin{figure}[t]
                  \centering        
                     \includegraphics[width=0.98\linewidth, clip=true, trim=0cm 9.7cm 0cm 0cm]{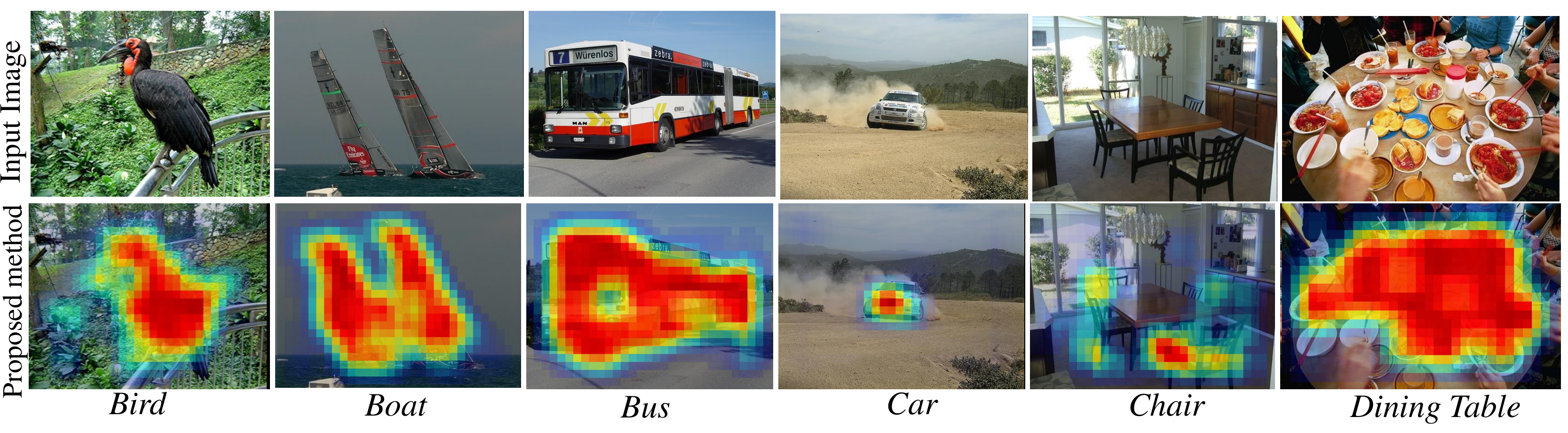}\\ 
                       \includegraphics[width=1\linewidth, clip=true, trim=0cm 11.2cm 0cm 0cm]{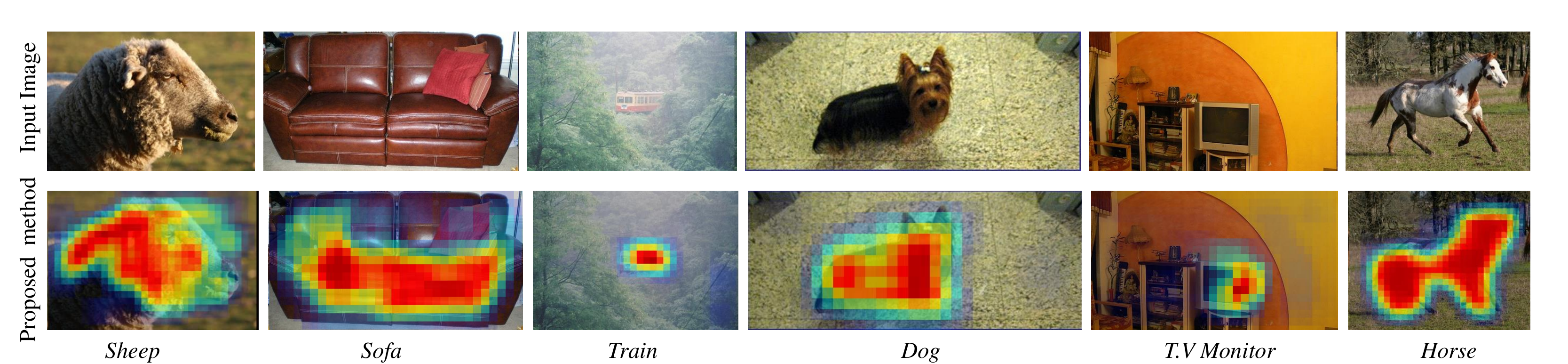}\\
                  \caption{Saliency detection by the proposed method on PASCAL VOC-07.}
                \label{fig:SalClassResults}
                        \end{figure}

                         Objects that are not visible clearly as in the \emph{Train} image and those in the presence of clutter as in \emph{TV Monitor} are  correctly detected in Fig.~\ref{fig:SalClassResults}. Even when a textureless  object such as Sofa, is large and occupies almost the entire image, our method shows good accuracy in marking the area as salient.
      \subsubsection*{ Images with multiple objects}
            \begin{figure*}[t]
                 \centering
                 \includegraphics[width=0.99\textwidth, clip=true, trim=0.4cm 10.7cm 13.5cm 0.2cm]{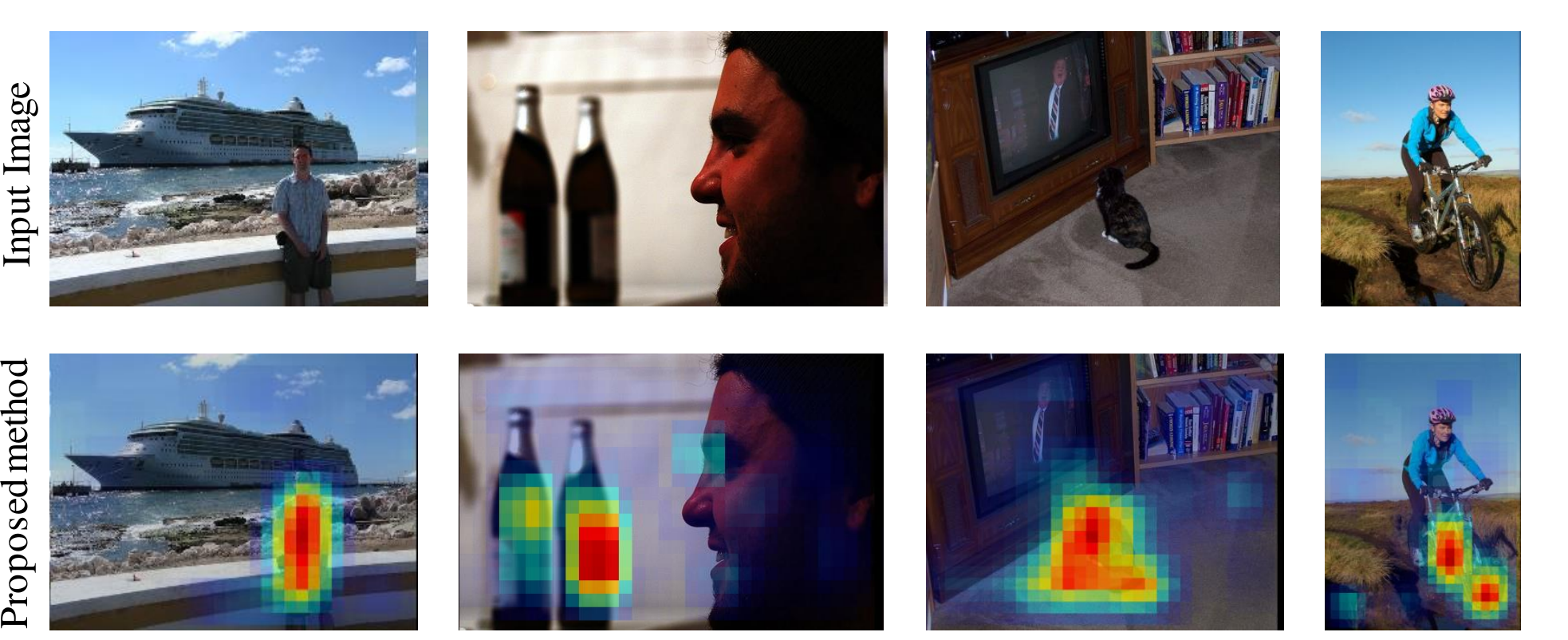}\\
                     \hspace{1.1cm} \textit{Person} \hspace{2.5cm} \textit{Bottle}   \hspace{2.9cm} \textit{Cat} \hspace{1.7cm} \textit{Bicycle}
                        \caption{Images with multiple objects from different categories.}
                 \label{fig:multipleCategimg}
                 \end{figure*}
            \begin{figure*}[t]
                 \centering
                 \includegraphics[width=1\textwidth, clip=true, trim=0.0cm 15.0cm 10.0cm 0cm]{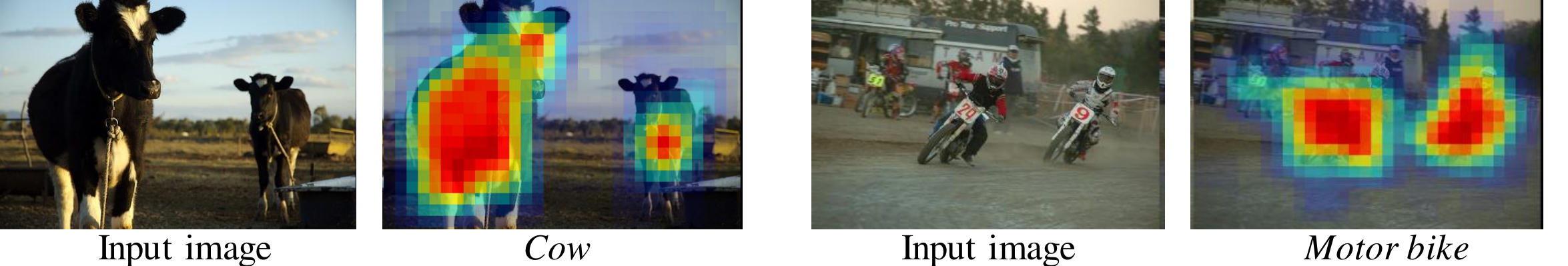}
                        \caption{Images with multiple objects from same category.}
                 \label{fig:multipleObjimg}
                 \end{figure*}

     Fig.~\ref{fig:multipleCategimg} shows the the ability of the proposed approach to discriminate among object categories on test images  in which objects from multiple categories are simultaneously present. These qualitative results are supported by the improved performance in   Fig.~\ref{fig:VOCResult}.  
 Even when a boat is present in the \emph{Person} image, the method marks only the person as salient. Similarly,  bottles and bicycle are correctly detected even though person is dominating in the \textit{bottle}, \textit{Bicycle}  images respectively.  
 The presence of dominant TV monitor, makes cat detection  a challenging task.
    Multiple instances of an object in  \emph{Cow} and \emph{Motorbike} have been successfully marked as salient in Fig.~\ref{fig:multipleObjimg}.

    \begin{figure*}[tp]
              \centering
       \includegraphics[width=1\linewidth, clip=true, trim=0cm 0cm 0cm 0cm]{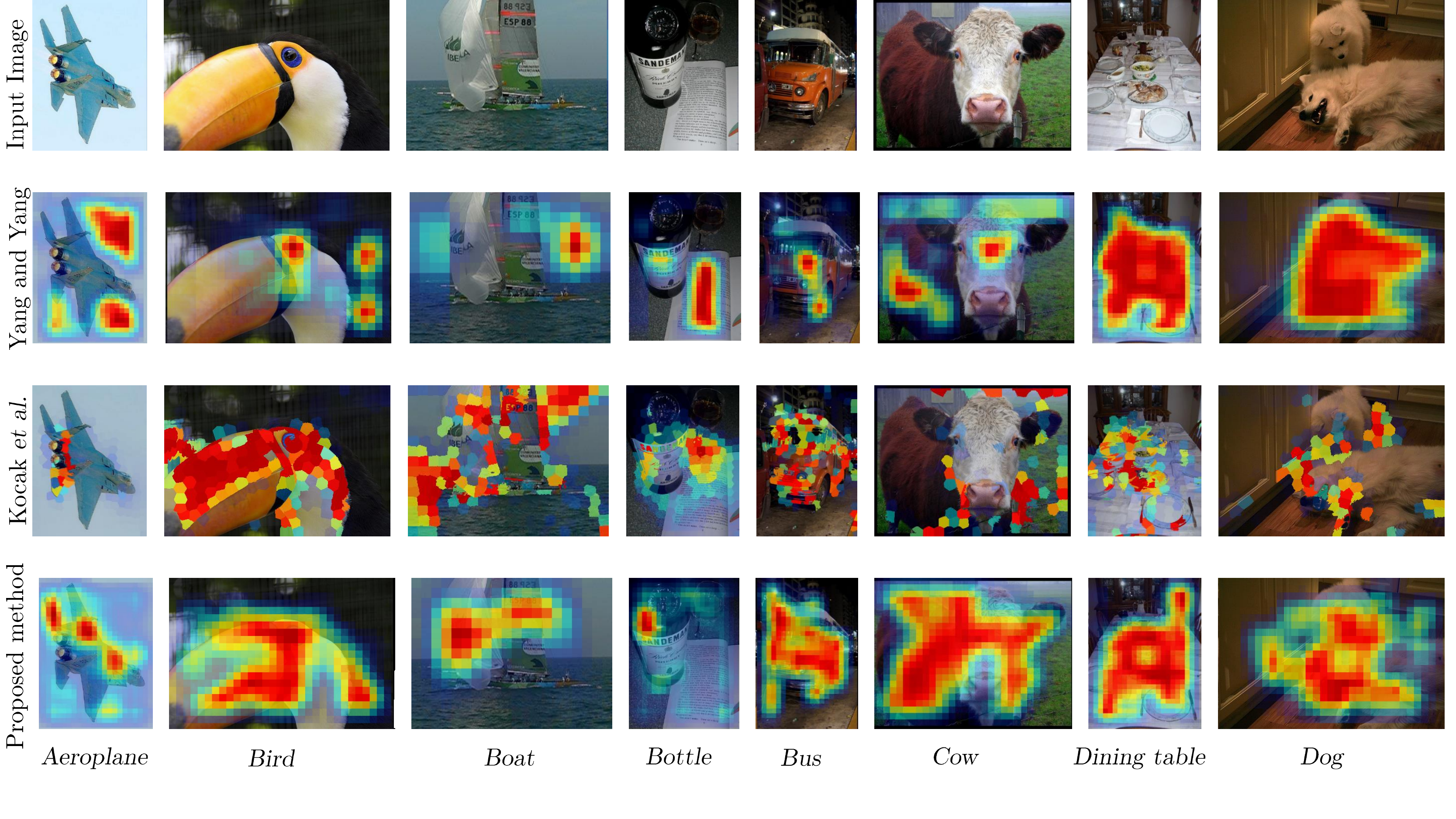}
       \includegraphics[width=1\linewidth, clip=true, trim=0cm 0cm 0cm 0cm]{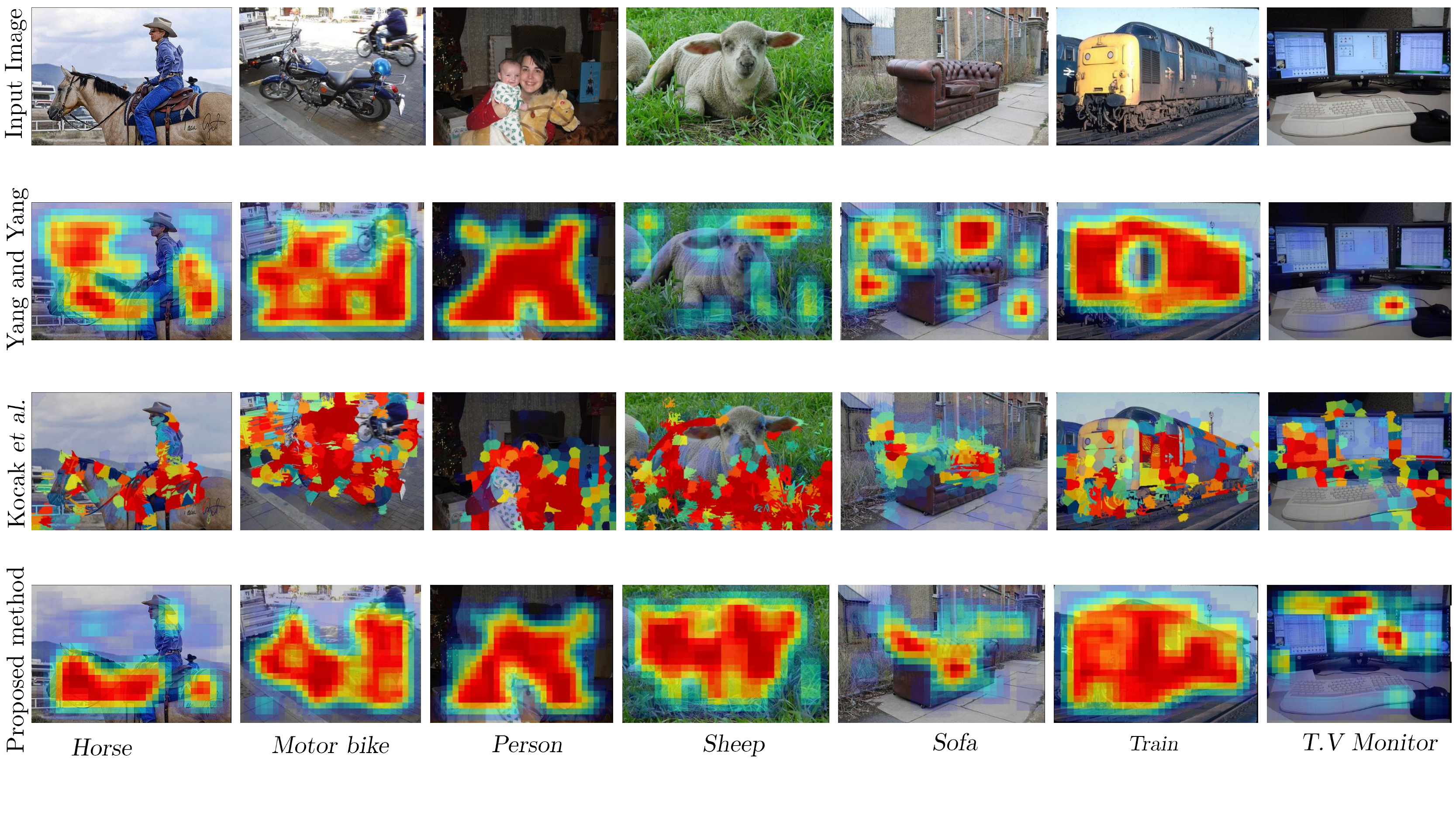}
       \caption{Qualitative comparison with Yang and Yang~\cite{topdownCVPR2012} and Kocak \textit{et al.}\cite{topdownBMVC2014} on PASCAL VOC-07 dataset}
       
       \label{fig:QualitativeComparisonYangVOC}
       \end{figure*}%
     
        \subsubsection*{ Qualitative comparison }
     
              Fig.~\ref{fig:QualitativeComparisonYangVOC} compares the saliency results of Yang and Yang~\cite{topdownCVPR2012} and Kocak \textit{et al.}~\cite{topdownBMVC2014} with our method. With the help of classifier-guided training, proposed method outperform ~\cite{topdownCVPR2012,topdownBMVC2014}, especially on test images in which object size is too big compared to the patch size as can be observed from  Fig.~\ref{fig:QualitativeComparisonYangVOC}~(\textit{Train}, \textit{Cow}). In accordance with the precision rates at EER results, Dog and Person saliency maps of~\cite{topdownCVPR2012}  are slightly better than proposed  method (see \textit{Dog} and ~ \textit{Person} in Fig.~\ref{fig:QualitativeComparisonYangVOC}). Improved Performance of proposed method on less textured object classes like aeroplane, bottle and sofa is clearly visible in Fig.~\ref{fig:QualitativeComparisonYangVOC}. 
   \subsection{Image segmentation}
      \begin{figure*}[t]
                    \centering
                    \includegraphics[width=1\textwidth, clip=true, trim=0.0cm 11.0cm 5.0cm 0cm]{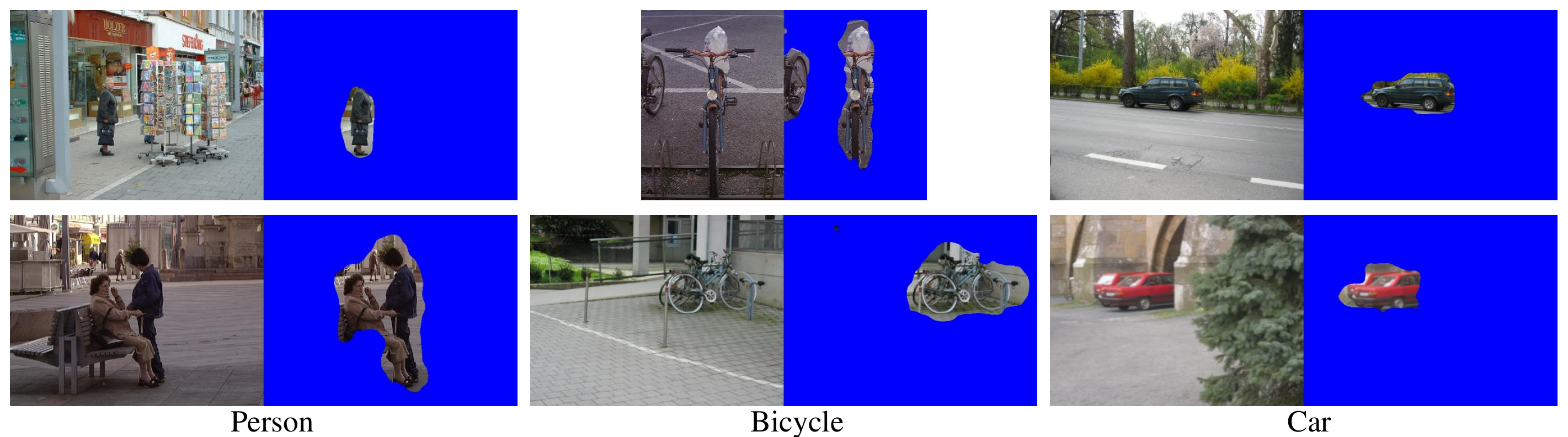}
                           \caption{Segmentation from the saliency map by simple thresholding.}
                    \label{fig:SegmentationGraz}
                    \end{figure*} 
    Although object segmentation in cluttered background is a challenging problem in itself, we investigate the effectiveness of using the saliency maps obtained by our method in segmentation.     
       The saliency map generated is a probability map with $1$ indicating the presence of object of interest and $0$ indicating absence of object. Here non-object includes background as well as  object pixels of negative classes. 
           The saliency models of all categories are  inferred  on a test image and  each pixel is assigned to the category of the saliency model that produces highest saliency value at that pixel. If the highest saliency at a pixel is below 0.5, those pixels are assigned to the background class.  Fig.~\ref{fig:SegmentationGraz} shows segmentation results achieved by this simple thresholding . Even though our  saliency models are trained and inferred at patch-level, it is capable of producing a segmentation that follows object boundaries. An improved pixel accurate segmentaion can be achieved by applying dedicated segmentation approaches such as Grabcut~\cite{rother2004grabcut}  on  salient regions. 
                            
        \subsection*{\textbf{Graz-02 dataset}}
        Intersection over union (IOU) is used as a  metric to evaluate segmentation performance, computed as $IOU = TP/(TP + FP + FN)$, where TP is number of true positive pixels, FP is the number of false positive pixels and FN is number of false negative
        pixels. Table \ref{table:Graz_SemanticSegInterSectUnion} compares the performance of the proposed method with \cite{globalBOFinCRFSeg2011} and \cite{JointSegCateg2012jain} on the Graz02 dataset. Following \cite{JointSegCateg2012jain}, the  proposed saliency models are evaluated on 150 test images from each object category~($3\times150$).
                
         Even with a simple thresholding of the saliency map, the segmentations produced with our method outperforms both the methods. 
          Better performance in background pixel classification illustrates the effect of classifier-weighted saliency inference, which reduced false positive detection.             
        \begin{table}[t]
         \centering
             \caption{Comparison with state-of-the-art semantic segmentation tasks on 450 test images of Graz-02 dataset using intersection over
             union metric. }
            \label{table:Graz_SemanticSegInterSectUnion}
      \resizebox{0.75\textwidth}{!}{
       \begin{tabular}{cccc}
       \hline
       Algorithm  & Singaraju and Vidal~\cite{globalBOFinCRFSeg2011} & Jain \textit{et al.}~\cite{JointSegCateg2012jain} & Proposed          \\ \hline
       Background & 82.32                         & 77.97                        & \textbf{83.46 }            \\
       Bike    & 46.18                         & \textbf{55.6 }                        & 50.02             \\
       Car        & 36.49                         & \textbf{41.51}                        & 40.81             \\
       Person     & \textbf{38.99}                         & 37.26                        & 38.8              \\ \midrule
       Mean       & 50.99                         & 53.08                        &\textbf{ 53.27}             \\ \hline
       \end{tabular}
       }
       \end{table}
        \subsection*{\textbf{PASCAL VOC-07 dataset}}
    Image segmentation approach \cite{aldavert2010fast} evaluates their model using pixel classification accuracy; i.e percentage of pixels correctly classified for each category. To compare with this approach, the proposed model  is also evaluated using this metric.
      The average classification accuracy (average accuracy of 21 categories- 20 object and one background category)  obtained by thresholding saliency maps at 0.5 is better (29.66\%) than image segmentation  approach~\cite{aldavert2010fast}~(23\%). Although our saliency maps are estimated using patch-level computations, the accuracy is  comparable to that of the superpixel based segmentation approach of \cite{fulkerson2009class}~( 27\% using 4 superpixel neighbors)
       \subsection{Image Classification}
              This paper proposes a novel and effective method for salient object detection in which the classifier plays an important role in improving results because of its ability to identify those objects whose models needs to be retrained by the CRF in order to remove false positives. Since there is a tight coupling between the saliency and classifier modules, it would be pertinent to investigate whether image classification performance gains with improved object models. 
       
\subsection*{\textbf{ Graz-02 dataset}} 
 \begin{table*}[t]
                            \centering
                            \caption{Classification accuracy for Graz-02.}
                            \label{table:Graz_classifier}
  \resizebox{0.9\textwidth}{!}{%
  \begin{tabular}{@{}llllll@{}}
  \toprule
  Algorithm & LLC~\cite{LLC_CVPR10} & Bilen \emph{et al}.~\cite{Vinay_IJCV}& ScSPM~\cite{ScSPM_CVPR09} & \begin{tabular}[c]{@{}l@{}}ScSPM using proposed\\  category-aware sparse coding \end{tabular} & \begin{tabular}[c]{@{}l@{}}Proposed saliency-weighted\\ classifier\end{tabular} \\ \midrule
  Accuracy & $87.8\pm 0.9$ & $91.18 \pm 1.4 $ & $88.5\pm 1$ & $89.5\pm 1.2$ &  $\textbf{91.21}\pm \textbf{1.2}$ \\ \bottomrule
  \end{tabular}
  }
  \end{table*}
 Table~\ref{table:Graz_classifier} compares the image classification accuracy of the classifier in the proposed framework
    with LLC~\cite{LLC_CVPR10}, ScSPM~\cite{ScSPM_CVPR09} and \cite{Vinay_IJCV}. In our implementation of ScSPM, we randomly sample 100 features from each training image and form a dictionary of size 2048. As suggested in \cite{LLC_CVPR10}, 5 local dictionary atoms are used to code a feature in LLC. Although LLC is faster than ScSPM, its classification accuracy is less.  When ScSPM is used in conjunction with our category-aware sparse coding, there is an improvement of about 1\%. However, when the saliency information is incorporated in the form of a weight derived from the classifier, the classification accuracy is 91.21\%, which is better than LLC and \cite{Vinay_IJCV}. It may be noted that in \cite{Vinay_IJCV}, the results are obtained using SIFT features extracted from multiple patch sizes for training the classifier using the AUC criterion. In \cite{segmentation_categ}, the authors achieve an accuracy of 92.23\% using multiple features such as color, shape and SIFT. Thus, although the classifier falls short of state-of-the-art performance by about 1\%, the top-down saliency framework provides a reasonably good classifier as an accessory.
\subsection*{\textbf{PASCAL VOC-07 dataset}}
In PASCAL VOC-07 dataset, objects from multiple categories are present in an image. So we train a binary SVM classifier for each object category as in \cite{LLC_CVPR10} instead of one-vs-rest SVM used for Graz-02 dataset. For example, in an image that contains both cat and TV monitor, the cat binary classifier will estimate the presence of cat in the image and  the TV monitor classifier will estimate the presence of  TV monitor in the image. If both classifier respond positively, the image will be marked as one containing both cat and TV monitor. On the other hand if the cat classifier wrongly estimated the absence of cat, then it is considered as a false negative for cat category. Average precision of each category is evaluated seperately on all  test images of PASCAL VOC-07 image classification dataset and the mean across 20 categories is evaluated.
      We achieved better mean average precision of 50.84\% as compared to 50.65\% by  ScSPM and better classification in 12 out of 20 object classes. Although the improvement is not significant, most of the errors are due to inter-class similarity, e.g. between bike and motor bike classes. The joint saliency-classifier framework needs $D_{sub}$ and a background dictionary in addition to the dictionaries for saliency modeling. These dictionaries are much smaller than the large dictionary~(12,000 atoms) used by ScSPM.              
       \subsection{Computation time}
       The training of joint framework for saliency estimation and image classification is  faster as compared to the saliency estimation approaches of \cite{topdownBMVC2014,topdownCVPR2012}, due to the  reduced number of iterations and images used by proposed classifier-guided training. MATLAB implementations of all approaches
         were evaluated on a PC running on Intel Xeon
              2.4GHz processor. Our training of all three object categories in Graz-02 and image classifier took just 3 hours and 34 minutes, while training of saliency models alone took 4 hours and 49 minutes by \cite{topdownCVPR2012} and 30 hours and 10 minutes by \cite{topdownBMVC2014}.
              
              The saliency estimation for 3  categories as well classifying the image took just 9.89 seconds. The SIFT feature extraction took 2.83 seconds, saliency inference per category took 1.72 seconds and category-aware sparse coding, saliency-weighted max-pooling, image classification  and saliency refinement took additional 1.9 seconds, totaling to 9.89 seconds ($3\times 1.72+2.83+1.9=9.9)$ seconds. The saliency estimation of 3 object categories alone took 7.99  seconds in \cite{topdownCVPR2012}.  The inference time is much larger in \cite{topdownBMVC2014}, which took 52 seconds. The ScSPM image classfication of \cite{ScSPM_CVPR09} alone need  6.9 seconds, due to conventional sparse coding on $D$, while with the help of category-aware sparse coding, the additional computational time for classification reduces to  1.9 seconds only . 
              For comparison, cascaded saliency and classification modules of \cite{topdownCVPR2012} and 
              \cite{ScSPM_CVPR09} requires 14.89 seconds per image, while our joint approach gives better accuracy in  9.89 seconds. 
              
%
%
%
%
%
%
%
%
           \section{Conclusion}
  In this work we propose  a framework for top-down salient object detection. Since the pipeline of image classification~\cite{ScSPM_CVPR09} and top-down saliency~\cite{topdownCVPR2012}  contains many common stages, our interconnected and mutually benefiting saliency-classification framework  reduces  the computational cost  compared to their independent implementations.  The image classifier is trained on novel category-aware sparse codes computed on  object dictionaries  used for saliency modeling. A novel  saliency-weighted max-pooling is proposed to improve image classification by weighting  the max-pooled vector  in each block of the spatial pyramid with a weight derived from net saliency of that block. Similarly, saliency maps are improved by using the image classifier that leverages  information about presence of the object  in an image.
 In the current implementation we extracted SIFT features using a fixed patch size, that reduces the performance of the model, if the object size is too small compared to the patch size. In the future,  we will implement a multiple-patch size, multi-feature based joint framework for saliency estimation and image classification.

%
%
  
\section{References}
 \bibliographystyle{elsarticle-num} 
  \bibliography{assoc}
\end{document}